\documentclass[manuscript,screen,nonacm]{acmart}

\usepackage{amsmath}
\usepackage{mathtools}
\usepackage{graphicx}
\usepackage{subcaption}
\usepackage{booktabs}
\usepackage{placeins}
   
\newcommand{\vect}[1]{\boldsymbol{#1}}
\newcommand{\mat}[1]{\mathbf{#1}}
\newcommand{\bM}{\mat{M}}
\newcommand{\bR}{\mat{R}}
\newcommand{\bU}{\mat{U}}
\newcommand{\bI}{\mat{I}}
\newcommand{\bK}{\mat{K}}
\newcommand{\bL}{\mat{L}}
\newcommand{\bD}{\mat{D}}
\newcommand{\bS}{\mat{S}}
\newcommand{\bX}{\mat{X}}
\newcommand{\bSigma}{\mat{\Sigma}}
\newcommand{\bs}{\vect{s}}
\newcommand{\bx}{\vect{x}}
\newcommand{\ba}{\vect{a}}
\newcommand{\bell}{\vect{\ell}}
\newcommand{\bphi}{\vect{\phi}}
\newcommand{\by}{\vect{y}}
\newcommand{\bv}{\vect{v}}
\newcommand{\bh}{\vect{h}}

\title{Interpretable Machine Learning for Spatial Science: A Lie-Algebraic Kernel for Rotationally Anisotropic Gaussian Processes}

\author{Kane Warrior}
\affiliation{%
  \institution{University of York}
  \department{Department of Mathematics}
  \city{York}
  \country{United Kingdom}
}
\email{gtq520@york.ac.uk}

\author{Dalia Chakrabarty}
\affiliation{%
  \institution{University of York}
  \department{Department of Mathematics}
  \city{York}
  \country{United Kingdom}
}
\email{dalia.chakrabarty@york.ac.uk}

\begin{abstract}
Many three-dimensional spatial fields are anisotropic, with directions of rapid and slow variation that need not align with the coordinate axes. Standard Gaussian process kernels with Automatic Relevance Determination (ARD) capture only axis-aligned anisotropy, while generic full symmetric positive definite (SPD) metrics can represent rotated anisotropy but do not parameterise principal length-scales and directions directly. We introduce an interpretable rotationally anisotropic GP kernel that parameterises a three-dimensional SPD covariance metric using three principal length-scales and an explicit $\mathrm{SO}(3)$ rotation. The rotation is represented by an axis--angle vector and mapped to $\mathrm{SO}(3)$ via the Lie-algebra exponential map, giving unconstrained Euclidean coordinates for inference while always inducing a valid SPD metric.

The construction spans the same family of three-dimensional SPD covariance metrics as a generic full-SPD parameterisation, but exposes the geometry differently: length-scales and orientation are explicit, interpretable, and directly available for prior specification and posterior summaries. We perform Bayesian inference on these quantities using Markov Chain Monte Carlo (MCMC), and characterise the resulting symmetries and weakly identified regimes.

On synthetic data with rotated anisotropy, the posterior recovers the generating metric and improves prediction relative to an axis-aligned ARD baseline, while matching the predictive performance of a generic full SPD baseline. When the ground truth is axis-aligned, posterior mass concentrates near the identity rotation and predictive performance matches ARD. On a material-density dataset from a laboratory-fabricated nano-brick, the inferred metric reveals rotated anisotropy that is not captured by axis-aligned kernels.

\end{abstract}


\begin{document}
\maketitle

\section{Introduction}

Across machine learning and artificial intelligence, there is increasing emphasis on models that are not only predictive, but also interpretable in terms of meaningful quantities, rather than relying only on post hoc explanations of black-box predictions \cite{rudin2019stop,murdoch2019definitions,rudin2022interpretable}. This is especially important in spatial and scientific applications, where the geometry of dependence can itself be of interest. Gaussian processes (GPs) are attractive in this setting because they provide flexible nonparametric regression with calibrated predictive uncertainty \cite{rasmussen2006gaussian}. However, the interpretability of a GP spatial model depends strongly on how the covariance structure is parameterised.

Many spatial fields are anisotropic. Correlations decay faster in some directions than others, and the directions of fast and slow variation need not align with the coordinate axes \cite{cressie1993statistics,stein1999interpolation,gelfand2010handbook}. In geostatistics, environmental modelling, and three-dimensional physical simulations \cite{cressie1993statistics,banerjee2014hierarchical,gelfand2010handbook}, the goal is therefore not only accurate prediction, but also an interpretable summary of the dominant directions and length-scales of spatial dependence \cite{zimmerman1993another,ecker1999bayesian}. In this paper, we address this need by parameterising GP anisotropy directly through principal length-scales and their orientation, so that the fitted covariance structure is interpretable as part of the model rather than only after fitting.

A common way to introduce anisotropy in GP kernels is automatic relevance determination (ARD), which assigns a separate length-scale to each input dimension \cite{neal1996bayesian,rasmussen2006gaussian}. ARD kernels are positive definite by construction and easy to interpret when the principal directions coincide with the coordinate axes. However, they remain axis-aligned. If the field varies most rapidly along a tilted direction in $\mathbb{R}^3$, ARD can only approximate this through a compromise in axis-aligned length-scales, which can blur the implied geometry of the anisotropy; see Section~\ref{sec:background-ard}.

A more flexible alternative is to use a full symmetric positive definite (SPD) metric $\bM\succ 0$, so that the covariance depends on quadratic forms $(\bx-\bx')^\top \bM(\bx-\bx')$ \cite{stein1999interpolation}. In three dimensions, $\bM$ has six degrees of freedom, and its eigendecomposition gives the principal directions and inverse squared length-scales of the covariance geometry \cite{zimmerman1993another,ecker1999bayesian,schmidt2003bayesian}. Generic SPD parameterisations, such as Cholesky-based parameterisations, are valid and expressive, but their free parameters are not themselves the principal length-scales and directions. As a result, the free parameters in a generic SPD parameterisation can have highly nonlinear and indirect relationships with the physical geometry of anisotropy. Geometric interpretation is therefore typically obtained only after fitting, through a post hoc eigendecomposition of $\bM$ \cite{kazianka2013objective}.

We address this by working directly with the two objects that define anisotropy: the three principal length-scales and their orientation. We use the structured metric
\begin{equation}
  \bM(\bell,\ba)
  =
  \bR(\ba)^\top
  \operatorname{diag}(\ell_x^{-2}, \ell_y^{-2}, \ell_z^{-2})
  \bR(\ba),
  \label{eq:intro-metric}
\end{equation}
where $\bell=(\ell_x,\ell_y,\ell_z)^\top$ specifies the principal length-scales and $\bR(\ba)$ rotates the corresponding principal directions into the original coordinate system. We parameterise $\bR(\ba)$ with an axis--angle vector $\ba\in\mathbb{R}^3$ via the Lie-algebra exponential map, so any $\ba\in\mathbb{R}^3$ maps to a valid rotation matrix and hence an SPD metric for every parameter value \cite{murray1994mathematical}. This separates the parameters controlling length-scale ($\bell$) from those controlling orientation ($\ba$). The special case $\ba=\vect{0}$ gives $\bR(\ba)=\bI$ and recovers the axis-aligned ARD metric as a nested baseline.

The structured metric $\bM(\bell,\ba)$ can be used inside standard stationary kernels such as the squared exponential and Mat\'ern families \cite{rasmussen2006gaussian,stein1999interpolation}. In terms of metric expressivity, this parameterisation has the same capacity as a generic full SPD metric in three dimensions. Its advantage is therefore not a larger covariance family, but a different statistical parameterisation: length-scales and orientation are explicit, interpretable, and directly available for prior specification, inference proposals, and posterior summaries. This is important because the nonlinear structure of the covariance geometry is separated into two meaningful components: correlation ranges, controlled by $\bell$, and orientation, controlled by $\ba$. Consequently, prior information about plausible correlation ranges or expected material directions can be encoded directly on these quantities, rather than indirectly through unconstrained entries of a generic SPD matrix. The same separation also makes posterior learning easier to diagnose, since uncertainty, weak identifiability, and symmetry-related behaviour can be studied in terms of ranges and rotations rather than opaque matrix entries. We infer $(\bell,\ba)$ with Markov Chain Monte Carlo (MCMC) under priors placed on these geometric quantities. On rotated synthetic data generated from the same kernel class, the model recovers the principal length-scales and orientation, while yielding lower test error than an axis-aligned ARD baseline. As expected, predictive performance is comparable to a generic full-SPD baseline, while giving more direct posterior summaries of length-scales and orientation. When the data are axis-aligned, posterior mass concentrates near the identity rotation, so the kernel effectively reduces to ARD. Implementation on a real dataset---that includes information of the value of the material density function at distinct 3D sub-surface locations---results in learning the metric that produces stable principal directions and length-scales, revealing the rotated anisotropy, which is, however, not captured by axis-aligned kernels.

Our contributions include: (i) a structured anisotropic GP kernel in $\mathbb{R}^3$ that parameterises the covariance metric directly in terms of three principal length-scales and an explicit rotation in $\mathrm{SO}(3)$; (ii) an axis--angle parameterisation, via the Lie algebra of $\mathrm{SO}(3)$, that allows unconstrained rotation coordinates during inference, while ensuring $\bR(\ba)\in \mathrm{SO}(3)$ and guaranteeing positive definiteness of the induced metric; (iii) an explicit relationship between the proposed parameterisation, ARD, and a generic full SPD metric: ARD is recovered at $\ba=\vect{0}$, while the full SPD baseline has comparable expressive power but less directly interpretable parameters. This distinction is important because our goal is not to enlarge the class of metrics beyond a generic full SPD parameterisation, but to make the same covariance geometry statistically interpretable: priors, proposals, diagnostics, and posterior summaries are placed directly on principal length-scales and orientation, rather than on matrix entries whose relationship to spatial structure is indirect and highly nonlinear.

The remainder of the paper reviews anisotropic GP kernels, develops the axis--angle parameterisation, describes MCMC inference for the structured metric, and compares the model with ARD and full-SPD baselines on synthetic and real three-dimensional data.

\section{Background and Related Work}

\subsection{Gaussian Processes and anisotropic kernels}
\label{sec:background-ard}

We briefly recall GP regression and the form of anisotropy considered in this paper. Let $f:\mathbb{R}^3\to\mathbb{R}$ be assigned a GP prior $f\sim\mathcal{GP}(0,k)$, and let the output variable realised at input $\bx=(x_1,x_2,x_3)^\top$ be $Y=f(\bx)$ \cite{rasmussen2006gaussian}. For stationary kernels, covariance depends on the difference $\bh=\bx-\bx'$, often through a quadratic form $\bh^\top \bM \bh$ with an SPD matrix $\bM$ \cite{stein1999interpolation}. The matrix $\bM$ determines the geometry of the covariance: its eigenvectors describe principal directions of correlation, while its eigenvalues determine inverse squared correlation ranges along those directions.

Automatic Relevance Determination (ARD) is a simple and widely used special case in which $\bM$ is diagonal. For example, the squared exponential ARD kernel can be written as
\begin{equation}
  k_{\mathrm{ARD}}(\bx,\bx')
  =
  \exp\!\Bigl(
    -\tfrac12 \sum_{j=1}^3 \frac{(x_j-x'_j)^2}{\ell_j^2}
  \Bigr),
  \label{eq:se-profile}
\end{equation}
with one length-scale per input dimension. ARD kernels are standard within machine learning \cite{neal1996bayesian,rasmussen2006gaussian} and appear throughout spatial statistics \cite{banerjee2014hierarchical}. The limitation for our setting is that ARD ties anisotropy to the coordinate axes: genuinely rotated structure in $\mathbb{R}^3$ cannot be represented directly, and must instead be approximated through a combination of axis-aligned length-scales.

More general anisotropic kernels can be obtained by using a full SPD matrix in the kernel distance. This is flexible, since a $3\times3$ SPD matrix has six degrees of freedom, but the free entries of a generic SPD parameterisation are not themselves the quantities one usually wants to interpret. For spatial applications, the scientifically meaningful quantities are typically the principal correlation ranges and their directions \cite{zimmerman1993another,ecker1999bayesian,schmidt2003bayesian}. Thus, while generic SPD parameterisations are valid for inference, their connection to the geometric expression of anisotropy is indirect \cite{kazianka2013objective}.

Several previous works have considered rotations in connection with Gaussian process models. One line of work uses feature-space rotations for Gaussian process classification, applying rotations to pairs of input dimensions so that generic feature vectors better match an isotropic GP classifier \cite{wang2012improved}. This setting is classification rather than spatial regression, and the rotation is used as a feature transformation rather than as an explicit spatial SPD metric. A more closely related construction introduces an anisotropic GP for traffic-state estimation using a rotated two-dimensional space--time kernel \cite{wu2024traffic}. That construction uses a single rotation angle in a $2\times2$ metric, with the angle interpreted in terms of traffic-wave propagation. In contrast, our work considers a generic three-dimensional spatial domain and parameterises a full $3\times3$ SPD metric in terms of three principal ranges and a full $\mathrm{SO}(3)$ orientation. We also perform Bayesian posterior inference over the ranges and orientation, and explicitly discuss the resulting symmetry and weak-identifiability structure.

\subsection{Rotations and Lie algebras in three dimensions}
\label{sec:so3-mat}

Rotations in three dimensions form the Lie group $\mathrm{SO}(3)$, with Lie algebra $\mathfrak{so}(3)$ given by $3\times3$ skew-symmetric matrices \cite{murray1994mathematical,gallier2020differential}. We use an axis--angle, or exponential-coordinate, parameterisation in which the rotation is indexed by a vector $\ba\in\mathbb{R}^3$. This vector maps to the skew-symmetric matrix
\begin{equation}
\label{eq:so3-A}
  \bU(\ba)
  =
  \begin{bmatrix}
    0 & -a_3 & a_2 \\
    a_3 & 0 & -a_1 \\
    -a_2 & a_1 & 0
  \end{bmatrix}
  \in \mathfrak{so}(3).
\end{equation}
The associated rotation matrix is obtained through the exponential map,
$\bR(\ba)=\exp(\bU(\ba))\in\mathrm{SO}(3)$, which can be evaluated using Rodrigues' formula \cite{murray1994mathematical,grassia1998practical}. Exponential coordinates are common in robotics, vision, and graphics because they provide a compact way to optimise or sample over rotations without imposing orthogonality constraints by hand \cite{murray1994mathematical,ma2004invitation,grassia1998practical}.

In GP modelling, anisotropy is more often represented either by axis-aligned ARD kernels or by generic SPD metrics \cite{stein1999interpolation,banerjee2014hierarchical}. The construction used here gives a full three-dimensional spatial metric with interpretable principal ranges, an explicit orientation, and a nested axis-aligned special case.

\section{Rotational Gaussian Process Kernel in 3D}
\label{sec:kernel}

We work with kernels on $\mathbb{R}^3$ of the form
\begin{equation}
  k(\bx,\bx')
  =  \kappa\!\Bigl(
    (\bx-\bx')^\top
    \bM(\bell,\ba)
    (\bx-\bx')
  \Bigr),
  \label{eq:kernel-metric}
\end{equation}
where $\kappa:[0,\infty)\to\mathbb{R}$ is a radial profile, e.g.\ squared exponential or Mat\'ern, and $\bM(\bell,\ba)$ is an SPD metric. The main modelling choice is how to parameterise $\bM$ so that anisotropy is both flexible and interpretable.

\subsection{Axis--angle coordinates via $\mathrm{SO}(3)$}

We use an explicit rotation $\bR(\ba)\in\mathrm{SO}(3)$, where $\mathrm{SO}(3)=\{\bR\in\mathbb{R}^{3\times3}:\bR^\top \bR=\bI,\ \det(\bR)=+1\}$, and parameterise it by an axis--angle vector $\ba\in\mathbb{R}^3$ through the Lie algebra $\mathfrak{so}(3)=\{\bU\in\mathbb{R}^{3\times3}:\bU^\top=-\bU\}$. We map $\ba$ to the skew-symmetric matrix $\bU(\ba)\in\mathfrak{so}(3)$ defined in Section~\ref{sec:so3-mat} in Eq.~\eqref{eq:so3-A}, and obtain the rotation via the exponential map
\begin{equation}
  \bR(\ba) := \exp\!\bigl(\bU(\ba)\bigr)\in\mathrm{SO}(3),
  \label{eq:R-of-a}
\end{equation}
which can be evaluated using Rodrigues' formula \cite{murray1994mathematical,grassia1998practical}. We treat $\ba$ as an unconstrained inference parameter in $\mathbb{R}^3$, and \eqref{eq:R-of-a} guarantees that every proposal induces a valid rotation matrix.

\subsection{Metric construction and induced kernel}

Given length-scales $\bell=(\ell_x,\ell_y,\ell_z)^\top$, define
$\Lambda(\bell):=\mathrm{diag}(\ell_x^{-2},\ell_y^{-2},\ell_z^{-2})$. We set
\begin{equation}
  \bM(\bell,\ba)
  := \bR(\ba)^\top \Lambda(\bell) \bR(\ba),
  \label{eq:metric-rotational}
\end{equation}
which is SPD whenever $\ell_x,\ell_y,\ell_z\neq 0$. Indeed, for any non-null $\bv$,
$\bv^\top \bM(\bell,\ba)\bv=(\bR(\ba)\bv)^\top\Lambda(\bell)(\bR(\ba)\bv)>0$.
Equivalently, the construction applies the linear transform
$\bx\mapsto \Lambda(\bell)^{1/2}\bR(\ba)\bx$ inside a radial kernel, so positive
definiteness follows directly. Substituting \eqref{eq:metric-rotational} into
\eqref{eq:kernel-metric}, the kernel depends on the squared rotated distance
$\psi(\bx,\bx')=(\bx-\bx')^\top\bM(\bell,\ba)(\bx-\bx')
=\|\Lambda(\bell)^{1/2}\bR(\ba)(\bx-\bx')\|_2^2$. The rotational kernel is then
\begin{equation}
  k_{\text{rot}}(\bx,\bx')
  =  \kappa\!\Bigl(\psi(\bx,\bx')\Bigr).
  \label{eq:rot-kernel}
\end{equation}
We consider $\kappa$ given either by the squared exponential profile
$\kappa_{\mathrm{SE}}(\psi)=\exp(-\psi/2)$ as in \eqref{eq:se-profile}, or by the
Mat\'ern family
\[
  \kappa_{\mathrm{Mat\acute{e}rn},\nu}(\psi)
  =
    \frac{1}{2^{\nu-1}\Gamma(\nu)}
    \bigl(\sqrt{2\nu\,\psi}\bigr)^{\nu}
    K_\nu\!\bigl(\sqrt{2\nu\,\psi}\bigr),
\]
with $K_\nu$ the modified Bessel function of the second kind. In this form, $\bell$
controls decay along the principal directions, while $\bR(\ba)$ sets their orientation
in the original coordinates.

\subsection{Identifiability and symmetry of $(\bell,\ba)$}
\label{subsec:identifiability}

Our kernel \eqref{eq:rot-kernel} is parameterised by principal ranges $\bell=(\ell_x,\ell_y,\ell_z)^\top$ and an orientation $\ba$ through $\bM(\bell,\ba)$ in \eqref{eq:metric-rotational}. The interpretability advantage of this parameterisation is that $\bell$ directly controls correlation decay along three orthogonal \emph{principal directions}, and $\bR(\ba)$ specifies how those directions are oriented in the original coordinates. Formally, the likelihood depends on $(\bell,\ba)$ only through the induced matrix $\bM$, so different parameter values can represent the same fitted anisotropy. We treat two parameter pairs as equivalent whenever they induce the same matrix, $(\bell,\ba) \sim (\bell',\ba') \iff \bM(\bell,\ba)=\bM(\bell',\ba')$. If $(\bell,\ba)\sim(\bell',\ba')$, then $(\bx-\bx')^\top \bM(\bx-\bx')$ is identical for all $\bx,\bx'$. Then kernel values, likelihoods, and posterior predictions coincide. This equivalence is standard when working with SPD matrices via their eigendecomposition and when using practical coordinates on $\mathrm{SO}(3)$ \cite{golub2013matrix,hartley2004multiple,mardia2000directional}. The key point for our model is that the data identify the anisotropic geometry encoded by $\bM$, while $(\bell,\ba)$ provide a structured and interpretable way to describe that geometry in terms of ranges and orientation, rather than through unconstrained entries of a generic SPD matrix.

The lack of identifiability, and the mitigation that we employ to address the same, are enumerated as follows. (i) When the three principal ranges are distinct, there are discrete symmetries: permuting the entries of $\bell$ and permuting the matching columns of $\bR(\ba)$ leaves $\bM=\bR(\ba)^\top\Lambda(\bell)\bR(\ba)$ unchanged, and eigenvector sign flips also leave $\bM$ unchanged. We impose a consistent labelling across draws via the eigendecomposition of $\bM$ (with ordered eigenvalues and using a fixed sign convention), while the parameterisation itself already provides the interpretable quantities: $\bell$ and $\bR(\ba)$. (ii) If $\ell_i\approx \ell_j$ for $i\neq j$ and $i,j\in\{x,y,z\}$, the orientation within the corresponding subspace is weakly identified: many rotations yield essentially the same matrix and the posterior for $\ba$ need not concentrate. This weak identification is itself informative, because it indicates that the data support similar correlation ranges in those directions and therefore do not strongly determine an orientation within the corresponding eigenspace. In a generic SPD parameterisation, the same phenomenon appears only indirectly through coupled movements of matrix entries or Cholesky parameters. In the proposed parameterisation, it is visible directly as uncertainty in orientation with little effect on the induced metric or predictions. Computationally, it also allows the chain to move in $\ba$ without forcing large coupled changes in $\bell$. (iii) As is standard for practical coordinates on $\mathrm{SO}(3)$, the axis--angle vector $\ba$ provides an unconstrained parameterisation for learning orientation while preserving a direct geometric meaning: its direction specifies the axis of rotation and its magnitude specifies the rotation angle. We therefore use $\ba$ for prior specification and inference, but report posterior summaries through the induced rotation matrix $\bR(\ba)$ and metric $\bM(\bell,\ba)$, using invariant quantities such as principal directions, principal ranges, and rotation angles. This is more interpretable than learning unconstrained entries of a generic SPD matrix, since the parameters being updated already correspond to the spatial geometry of anisotropy rather than to matrix entries whose physical meaning is only obtained after eigendecomposition.

\subsection{Why parameterise in the Lie algebra?}
\label{subsec:why-lie}

We use exponential coordinates in $\mathfrak{so}(3)$ rather than Euler angles because the axis--angle vector $\ba\in\mathbb{R}^3$ is an unconstrained Euclidean parameter, while the map $\ba\mapsto\bU(\ba)\mapsto\bR(\ba)$ always produces a valid rotation matrix in $\mathrm{SO}(3)$. This lets us place a Gaussian prior on $\ba$ and update it with standard random-walk proposals in MCMC without enforcing orthogonality constraints \cite{murray1994mathematical,barfoot2017state,gallier2020differential}. It also avoids the coordinate singularities and trigonometric coupling associated with Euler angles \cite{shuster1993survey,diebel2006attitude,grassia1998practical}. Thus anisotropy remains described by two interpretable objects: the principal ranges $\bell$ and their orientation $\bR(\ba)$.

\subsection{Inference}
\label{sec:inference}

We learn the random function $f(\cdot)$, where $\Phi=f(\bx)$, by modelling $f\sim\mathcal{GP}(0,k_{\vect{\theta}})$. Here $\bX$ is the three-dimensional spatial location vector, $\Phi\in\mathbb{R}$ is realised at $\bX=\bx$, and the model parameter vector is $\mat{\Theta}=\vect{\theta}$, with $\vect{\theta}=(\bell^\top,\ba^\top)^\top$. The vector $\bell=(\ell_x,\ell_y,\ell_z)^\top$ contains the range parameters, and $\ba\in\mathbb{R}^3$ specifies the orientation via $\bR(\ba)=\exp(\bU(\ba))$.

We learn $f$ using the training set $\bD=\{(\bx_i,\phi_i)\}_{i=1}^n$. For random variables $\Phi_i$ realised at the design points $\bx_i$, the joint distribution of $\Phi_1,\ldots,\Phi_n$ is multivariate normal with covariance matrix $\bK_{\vect{\theta}}\in\mathbb{R}^{n\times n}$, where $\bK_{\vect{\theta}}=[k_{\theta}(\bx_i,\bx_j)]$. The log-likelihood is
\begin{equation}
  \log \mathcal{L}(\vect{\theta})
  =
  -\tfrac12 \bphi^\top \bK_{\vect{\theta}}^{-1}\bphi
  -\tfrac12 \log\det \bK_{\vect{\theta}}
  -\tfrac{n}{2}\log(2\pi),
  \label{eq:loglik}
\end{equation}
where $\bphi=(\phi_1,\ldots,\phi_n)^\top$. We place independent Gaussian priors on the components of $\mat{\Theta}$, with $\ell_j\sim\mathcal{N}(\mu_{\ell_j},\sigma_{\ell_j}^2)$ for $j\in\{x,y,z\}$ and $\ba\sim\mathcal{N}(\vect{0},\sigma_a^2\bI_3)$. The prior on $\ba$ is centred at the identity rotation since we do not wish to impose strong prior information on orientation in a new application. Using $\pi_0(\vect{\theta})$ for the prior, the posterior is $\pi(\vect{\theta}\mid\bD)\propto \mathcal{L}(\bphi\mid\bx,\vect{\theta})\pi_0(\vect{\theta})$.

We sample from $\pi(\vect{\theta}\mid\bD)$ using random-walk Metropolis--Hastings. Proposals for the rotation coordinates $\ba\in\mathbb{R}^3$ are made in Euclidean space and mapped to $\bR(\ba)$ via \eqref{eq:R-of-a}, guaranteeing a valid rotation and hence an SPD metric $\bM(\bell,\ba)$ for every proposal with nonzero range parameters. We propose $\ell_i$ from a normal distribution with an experimentally chosen variance $\sigma_i$ for $i\in\{x,y,z\}$.

Learning $f(\cdot)$ as a sample function of a GP implies that the posterior predictive distribution of $\Phi_\ast$ at a new input $\bX=\bx_\ast$ has closed-form expectation and variance. The ARD baseline uses the same inference scheme with $\ba=\vect{0}$, so $\bR(\ba)=\bI$. We also compare with a generic SPD baseline that learns $\bM=\bL\bL^\top$, where $\bL$ is lower triangular with positive diagonal entries; the free entries of $\bL$ are updated using random-walk Metropolis--Hastings.

\section{Synthetic experiments}
\label{sec:synthetic}

We first study the behaviour of the rotational kernel on controlled three-dimensional synthetic datasets. The goals are to assess whether the model can recover known length-scales and rotation when the data are generated from a rotated anisotropic kernel, and to verify that the rotational kernel does not degrade performance when the ground-truth anisotropy is axis-aligned. Throughout, we compare the proposed rotational kernel against a standard axis-aligned ARD kernel and a generic SPD baseline, so that any gain from rotation can be separated from the more general effect of using a full SPD metric. All models use the same likelihood function and the same random-walk Metropolis--Hastings algorithm for inference. They differ only in how $\bM$ is parameterised: while the rotational kernel uses $(\bell,\ba)$ (Section~\ref{sec:kernel}), ARD restricts $\bM$ to be diagonal. The SPD baseline uses an unconstrained Cholesky parameterisation $\bM = \bL\bL^\top$ with six free parameters. These six parameters are the three unconstrained sub-diagonal entries of the lower-triangular factor $\bL$ together with its three strictly positive diagonal entries.

\subsection{Synthetic data generation}
\label{subsec:synthetic-gen}

We generate two synthetic datasets, denoted $\bD_1$ and $\bD_2$, using the same design-point construction and noise level, but with different ground-truth kernel orientations.

We draw $N_{\text{train}} = 1000$ training inputs and $N_{\text{test}} = 500$ test inputs, i.e.\ a total of $N_{\text{total}} = 1500$ inputs, independently from the uniform distribution on the cube $[-1,1]^3 \subset \mathbb{R}^3$. Let the full set of input locations be $ \bS_{\text{all}} := \{\bx_i\}_{i=1}^{N_{\text{total}}} \subset [-1,1]^3$, with $\bx_i \sim \mathcal{U}[-1,1]^3.$ We form training and test input sets by drawing a random permutation of index values in $\{1,\ldots,N_{\text{total}}\}$ and taking the first $N_{\text{train}}$ indices as training and the remaining $N_{\text{test}}$ indices as test, yielding $\bS_{\text{train}}$ and $\bS_{\text{test}}$ respectively. Then $\{\bx_i\}_{i\in \bS_{\text{train}}}$ are the design inputs while $\{\bx_i\}_{i\in \bS_{\text{test}}}$ are the test inputs.

Dataset $\bD_1$ is generated from a zero-mean Gaussian process with the rotational kernel $k_{\text{rot}}$ in Eq.~\eqref{eq:rot-kernel}, using $\bM(\bell,\ba)$ as defined in Eq.~\eqref{eq:metric-rotational}. For data generation we set
\[
  \bell_{\text{true}}=(0.40,\;0.10,\;0.80),
  \qquad
  \ba_{\text{true}}=(0.7,\;-0.4,\;1.0)^\top.
\]
Let $\bSigma^{(1)}_{\text{true}}$ denote the resulting $N_{\text{total}}\times N_{\text{total}}$ correlation matrix: $\bSigma^{(1)}_{\text{true}} = [\sigma_{i,j}]$. $\mathrm{MVN}(\vect{\mu},\bSigma)$ denotes the multivariate normal distribution with mean vector $\vect{\mu}$ and covariance matrix $\bSigma$. We then draw the vector of outputs, as realised at these inputs as: $\by^{(1)} := (y^{(1)}_1, \ldots, y^{(1)}_{N_{\text{total}}})^\top \sim \mathrm{MVN}(\vect{0}, \bSigma^{(1)}_{\text{true}} + \sigma_{\text{noise}}^2 \bI)$. The noise is considered Gaussian with standard deviation $\sigma_{\text{noise}}=0.05$.

Inputs marked by indices in $\bS_{\text{train}}$ and the corresponding components of $\by^{(1)}$ are used to populate a training set $\bD_1 =\{(\bx_i, y^{(1)}_i)\}_{i\in \bS_{\text{train}}}$.

Dataset $\bD_2$ is generated from a zero-mean Gaussian process, where the generating kernel matrix $\bSigma^{(2)}_{\text{true}}$ is constructed from the ARD kernel in Eq.~\eqref{eq:se-profile} with the true length-scale vector:
\[
\bell_{\text{true,axis}}=(1.00,\;0.25,\;0.37)^\top.
\]
Again, we have output vector $\by^{(2)}$, generated by $\by^{(2)}:=(y^{(2)}_1, \ldots, y^{(2)}_{N_{\text{total}}})^\top\sim\mathrm{MVN}(\vect{0},\bSigma^{(2)}_{\text{true}}+ \sigma_{\text{noise}}^2 \bI)$, with Gaussian noise with standard deviation $\sigma_{\text{noise}} = 0.05$. In the same way as $\bD_1$, inputs marked by indices in $\bS_{\text{train}}$ and the corresponding components of $\by^{(2)}$ are used to populate a training set $\bD_2 =\{(\bx_i, y^{(2)}_i)\}_{i\in \bS_{\text{train}}}$.

We learn three GP kernel parameterisations on both $\bD_1$ and $\bD_2$; the ARD baseline restricts $\bM$ to be diagonal; and the generic SPD baseline uses a Cholesky parameterisation $\bM=\bL\bL^\top$. In all cases we run a single MCMC chain for $100{,}000$ iterations with a burn-in of $50{,}000$, and compute posterior summaries from the post burn-in draws.

\subsection{Hyperparameter recovery and predictive performance}

We first look at the learnt hyperparameter recovery on the two synthetic datasets $\bD_1$ and $\bD_2$. Table~\ref{tab:synthetic-hyperparams} reports learnt hyperparameters for the rotational and ARD fits alongside the true data-generating values; for ARD the rotation parameters are not applicable.

\begin{table}[!htbp]
  \centering
  \caption{Synthetic-data hyperparameter recovery. Rotation parameters are not applicable for ARD.}
  \label{tab:synthetic-hyperparams}
  \small
  \setlength{\tabcolsep}{5pt}

  \begin{tabular}{lrrrrrr}
    \toprule
    \multicolumn{7}{c}{$\bD_1$: rotated anisotropy} \\
    \midrule
    Model
    & $\ell_x$ & $\ell_y$ & $\ell_z$ & $a_1$ & $a_2$ & $a_3$ \\
    \midrule
    True
    & 0.4000 & 0.1000 & 0.8000 & 0.7000 & -0.4000 & 1.0000 \\
    Rot.
    & 0.4057 & 0.0997 & 0.8009 & 0.6827 & -0.4403 & 1.0093 \\
    ARD
    & 0.1622 & 0.2280 & 0.1319 & -- & -- & -- \\
    \bottomrule
  \end{tabular}

  \vspace{0.8em}

  \begin{tabular}{lrrrrrr}
    \toprule
    \multicolumn{7}{c}{$\bD_2$: axis-aligned anisotropy} \\
    \midrule
    Model
    & $\ell_x$ & $\ell_y$ & $\ell_z$ & $a_1$ & $a_2$ & $a_3$ \\
    \midrule
    True
    & 1.0000 & 0.2500 & 0.3700 & 0.0000 & 0.0000 & 0.0000 \\
    Rot.
    & 0.9868 & 0.2510 & 0.3670 & -0.0317 & -0.0002 & -0.0011 \\
    ARD
    & 0.9854 & 0.2518 & 0.3668 & -- & -- & -- \\
    \bottomrule
  \end{tabular}
\end{table}

To report anisotropy in a form that does not depend on the raw rotation coordinates, we summarise posterior draws through the eigendecomposition of the induced matrix. For the rotated synthetic problem, the eigenvalues of $\bM_{\text{true}}$ are $\lambda_{\text{true}}=(100.00,\;6.25,\;1.5625)$, corresponding to principal correlation ranges $\tilde{\bell}_{\text{true}}=(0.10,\;0.40,\;0.80)$ (ordered from shortest to longest range). Using the learnt parameters from Table~\ref{tab:synthetic-hyperparams}, the learnt rotational matrix has eigenvalues $\lambda=(100.60,\;6.08,\;1.559)$ and hence principal ranges $\tilde{\bell}=(0.0997,\;0.4057,\;0.8009)$. The associated principal directions recover the generating orientation with eigenpairs ordered by eigenvalue and a deterministic eigenvector sign convention. The axis-wise misalignment angles $\theta_i=\arccos(|\hat q_i^\top q_{\text{true},i}|)$ are $(0.44^\circ,\;2.39^\circ,\;2.38^\circ)$.

For the generic SPD baseline, we apply the same eigendecomposition summary to $\bM=\bL\bL^\top$. On the rotated dataset the learnt SPD matrix has eigenvalues $\lambda_{\text{SPD}}=(100.67,\;6.19,\;1.553)$ and principal ranges $\tilde{\bell}_{\text{SPD}}=(0.0997,\;0.4019,\;0.8023)$. Comparing principal directions to the ground truth yields misalignment angles $(0.12^\circ,\;4.20^\circ,\;4.08^\circ)$. Thus the SPD baseline recovers essentially the same anisotropy as the rotational model. However, the rotational parameterisation exposes ranges and orientation directly, while the SPD parameterisation represents them only implicitly through the entries of $\bM$.

Given dataset $\bD_1$, the rotational kernel successfully recovers both the length-scales and the axis--angle vector to within sampling error, whereas the ARD baseline settles on a very different set of axis-aligned length-scales in an attempt to account for the rotated anisotropy. Given dataset $\bD_2$, both rotational and ARD recover length-scales close to the truth, and the learnt $\ba$ for the rotational kernel is near zero, indicating that the model effectively collapses back to a non-rotated matrix when appropriate. The SPD baseline similarly concentrates near an approximately diagonal matrix; in the axis-aligned case its learnt matrix has eigenvalues $\lambda_{\text{SPD}}=(16.00,\;7.46,\;1.02)$ corresponding to principal ranges $(0.2500,\;0.3662,\;0.9891)$, close to the ground-truth ranges $(0.25,\;0.37,\;1.00)$, with small direction misalignment $(1.75^\circ,\;1.75^\circ,\;0.10^\circ)$.
Table~\ref{tab:synthetic-pred} compares the performance of the three models on both datasets using mean absolute error (MAE), coverage of nominal $68\%$ and $95\%$ prediction intervals, and the standard deviation of standardised residuals.
\begin{table}[!htbp]
  \centering
  \caption{Synthetic test performance. Coverage uses nominal $68\%$ and $95\%$ predictive intervals.}
  \label{tab:synthetic-pred}

  \setlength{\tabcolsep}{8pt}

  \begin{tabular}{lcccc}
    \toprule
    \multicolumn{5}{c}{$\bD_1$: rotated anisotropy} \\
    \midrule
    Model & MAE & $\mathrm{cov}_{68}$ & $\mathrm{cov}_{95}$ & $\mathrm{std}(z)$ \\
    \midrule
    Rotational & 0.1252 & 0.698 & 0.954 & 0.964 \\
    SPD        & 0.1255 & 0.692 & 0.954 & 0.935 \\
    ARD        & 0.4709 & 0.678 & 0.934 & 0.864 \\
    \bottomrule
  \end{tabular}

  \vspace{0.8em}

  \begin{tabular}{lcccc}
    \toprule
    \multicolumn{5}{c}{$\bD_2$: axis-aligned anisotropy} \\
    \midrule
    Model & MAE & $\mathrm{cov}_{68}$ & $\mathrm{cov}_{95}$ & $\mathrm{std}(z)$ \\
    \midrule
    Rotational & 0.0656 & 0.672 & 0.940 & 1.029 \\
    SPD        & 0.0655 & 0.672 & 0.940 & 1.026 \\
    ARD        & 0.0656 & 0.672 & 0.940 & 1.025 \\
    \bottomrule
  \end{tabular}
\end{table}
In terms of prediction, the rotational and generic SPD kernels perform nearly identically on the rotated dataset $\bD_1$, whereas ARD has substantially larger error, consistent with its inability to represent rotated anisotropy directly. On the axis-aligned dataset $\bD_2$, all three models perform similarly, indicating that allowing rotation does not degrade performance when the true orientation is axis-aligned.

\subsection{Posterior predictive visualisations}
\label{subsec:synthetic-pred-plots}

To complement the quantitative results in Table~\ref{tab:synthetic-pred}, we visualise posterior predictions on both synthetic datasets. These plots compare the axis-aligned ARD baseline with the proposed rotational kernel and illustrate how the learnt covariance geometry affects prediction.

\begin{figure}[!htbp]
  \centering

  \begin{minipage}{0.49\columnwidth}
    \centering
    \includegraphics[width=\linewidth]{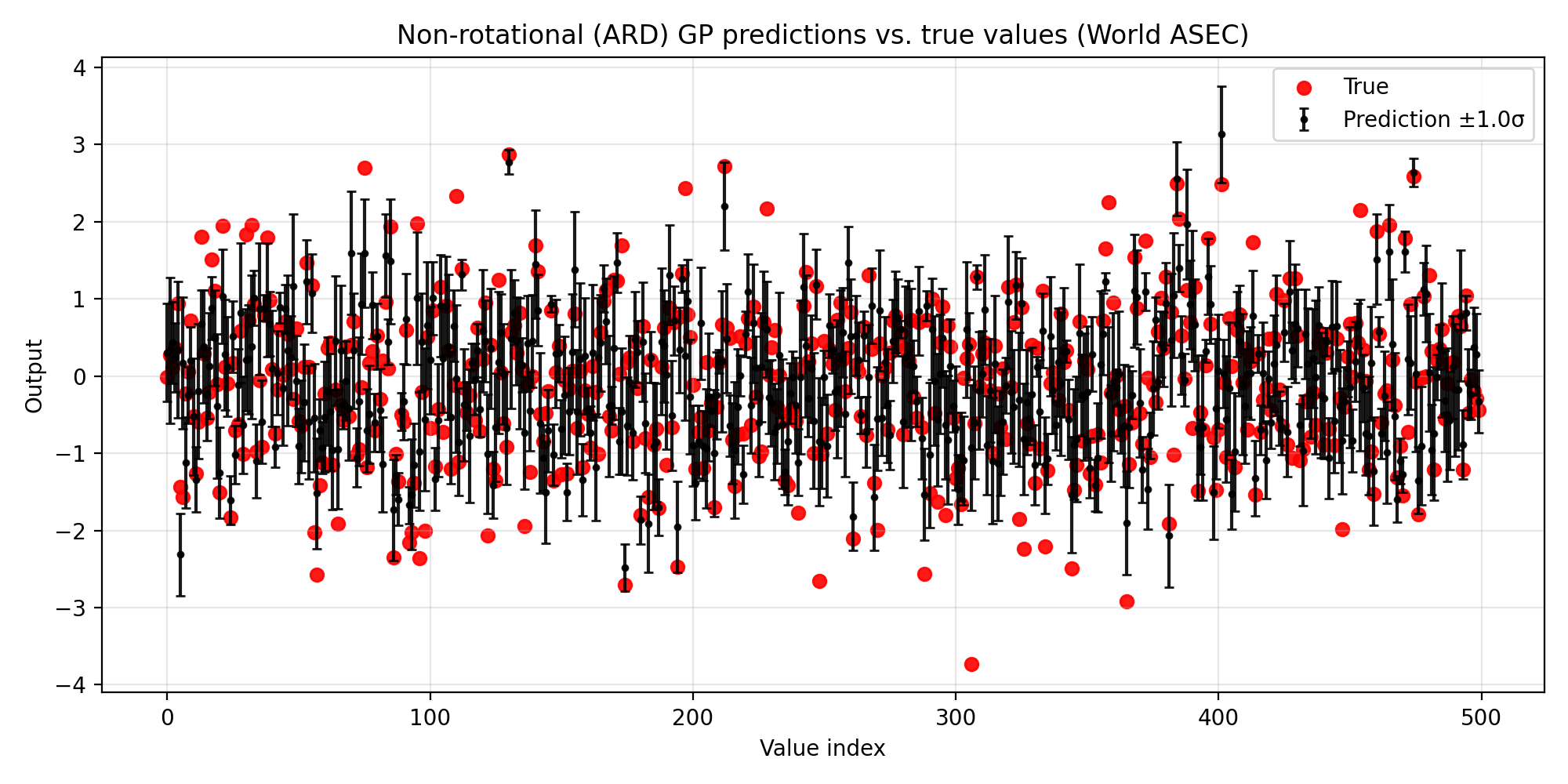}

    \vspace{0.4em}

    \includegraphics[width=\linewidth]{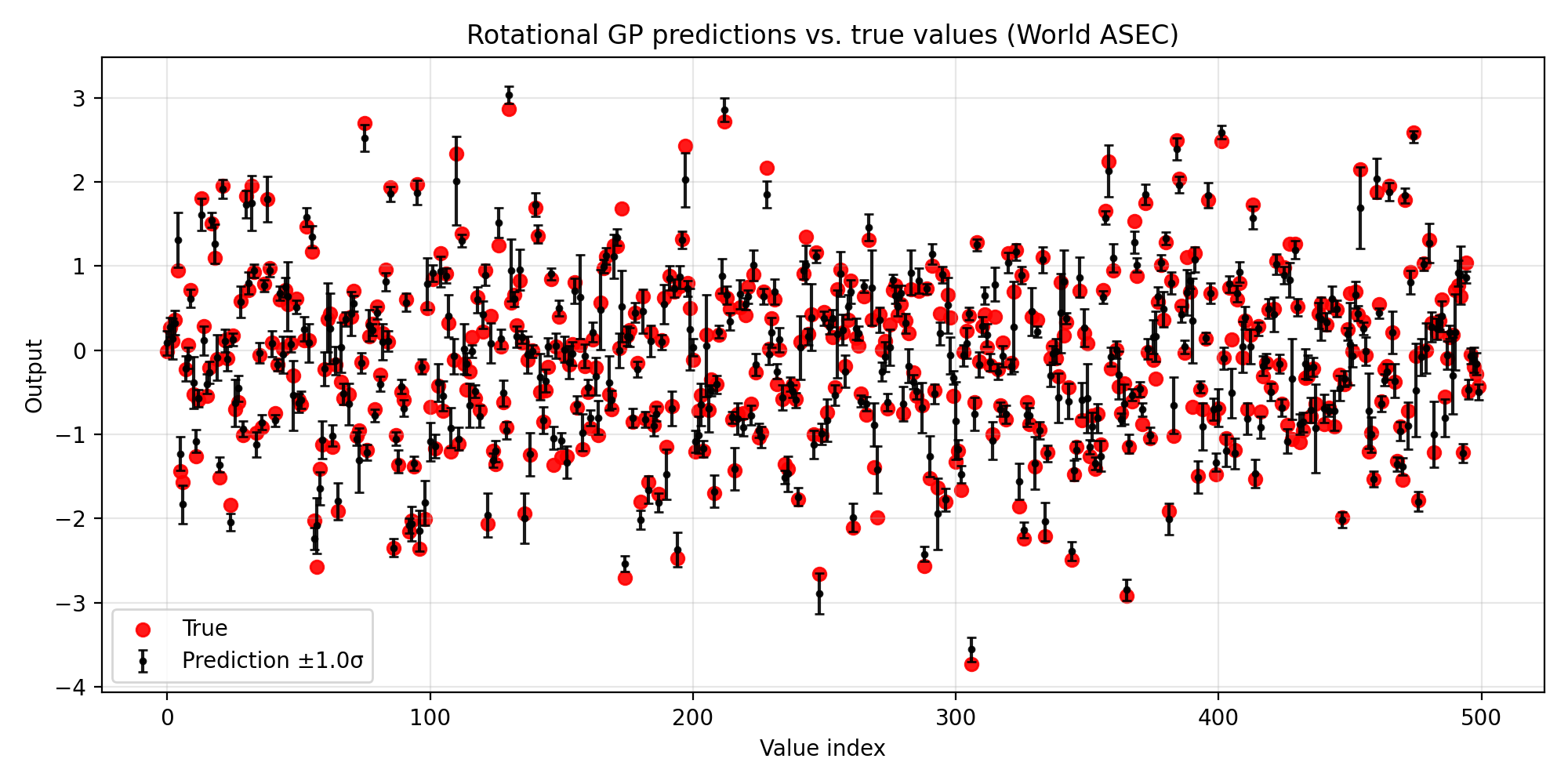}

    \vspace{0.2em}
    {\small $\bD_1$: rotated synthetic data}
  \end{minipage}
  \hfill
  \begin{minipage}{0.49\columnwidth}
    \centering
    \includegraphics[width=\linewidth]{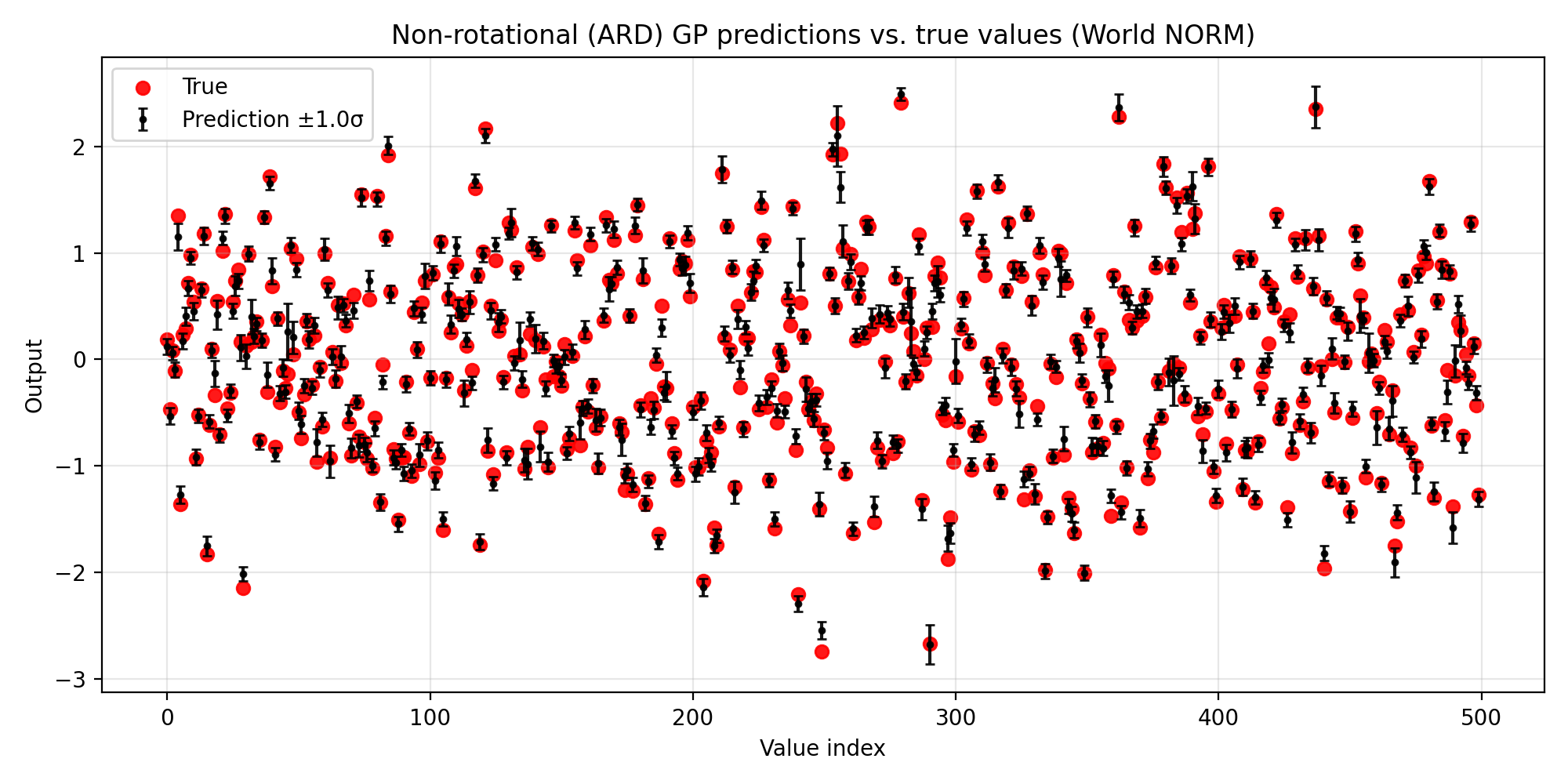}

    \vspace{0.4em}

    \includegraphics[width=\linewidth]{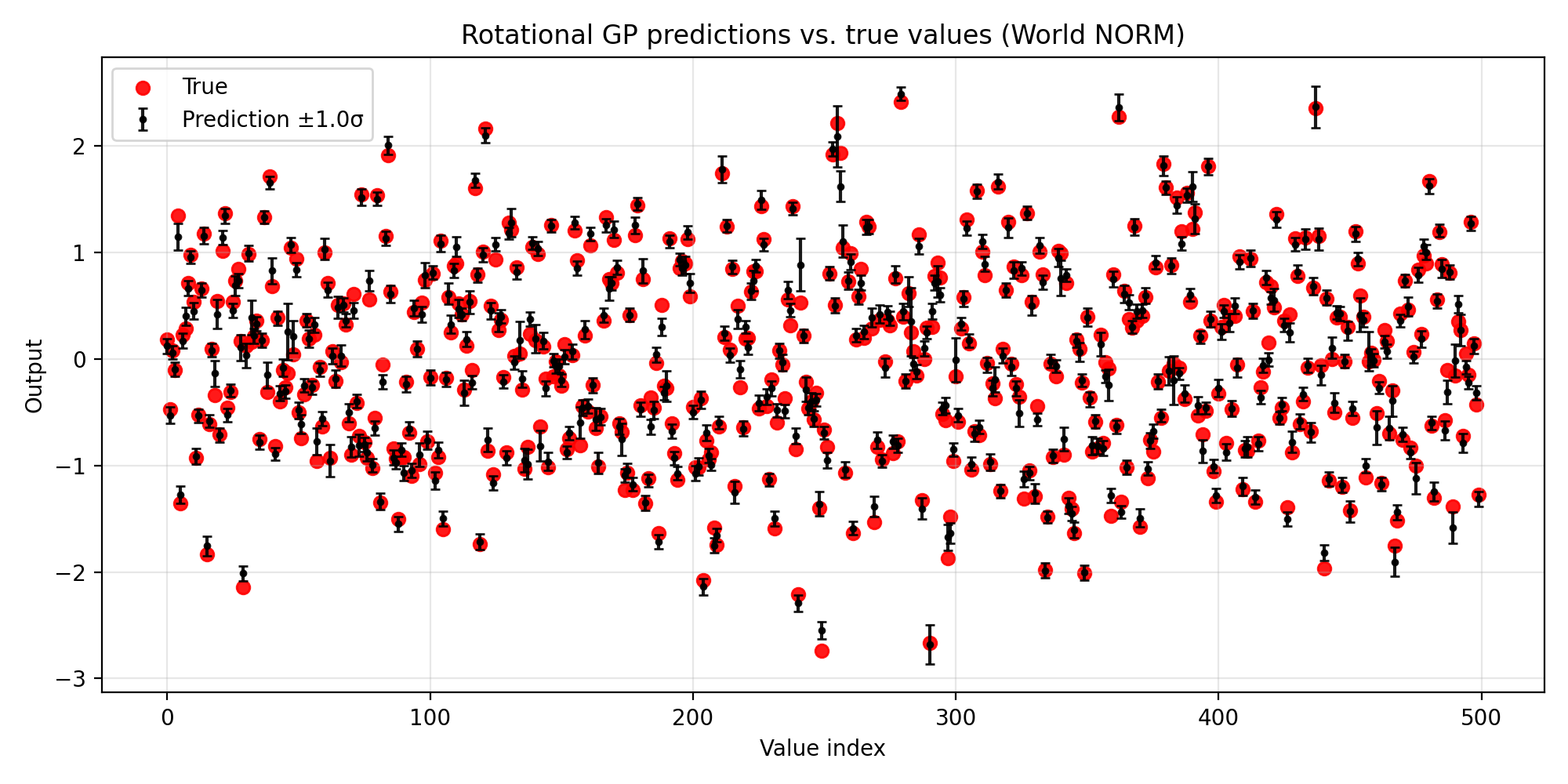}

    \vspace{0.2em}
    {\small $\bD_2$: axis-aligned synthetic data}
  \end{minipage}

  \caption{Posterior predictions on the synthetic datasets. In each column, the top panel shows axis-aligned ARD kernel predictions and the bottom panel shows rotational-kernel predictions. Error bars show $\pm1\sigma$ predictive uncertainty from the closed-form predictive variance.}
  \label{fig:pred-synthetic-both}
\end{figure}

Figure~\ref{fig:pred-synthetic-both} shows predictions on both synthetic datasets. For the rotated synthetic dataset $\bD_1$, the ground-truth anisotropy is rotated, so the axis-aligned ARD model cannot represent the correlation structure directly and gives visibly poorer predictions. By contrast, the rotational kernel captures the rotated anisotropy and produces predictions that better match the held-out data. For the axis-aligned synthetic dataset $\bD_2$, which was generated from an axis-aligned ARD kernel, the rotational kernel does not degrade performance: its posterior concentrates near an approximately axis-aligned metric, yielding predictions comparable to the axis-aligned baseline.

\FloatBarrier

\subsection{Trace diagnostics}
\label{subsec:synthetic-traces}

\begin{figure}[!htbp]
  \centering
  \begin{minipage}{0.65\columnwidth}
    \centering
    \includegraphics[width=\linewidth]{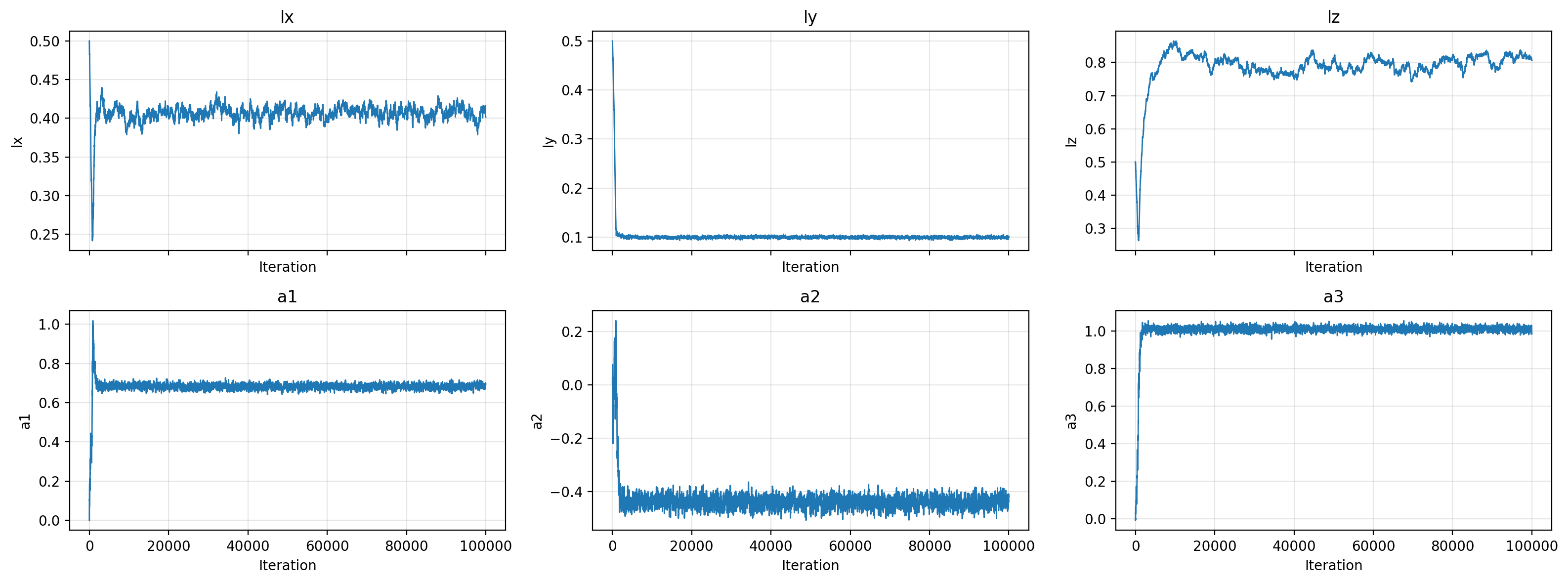}
  \end{minipage}

  \vspace{0.5em}

  \begin{minipage}{0.65\columnwidth}
    \centering
    \includegraphics[width=\linewidth]{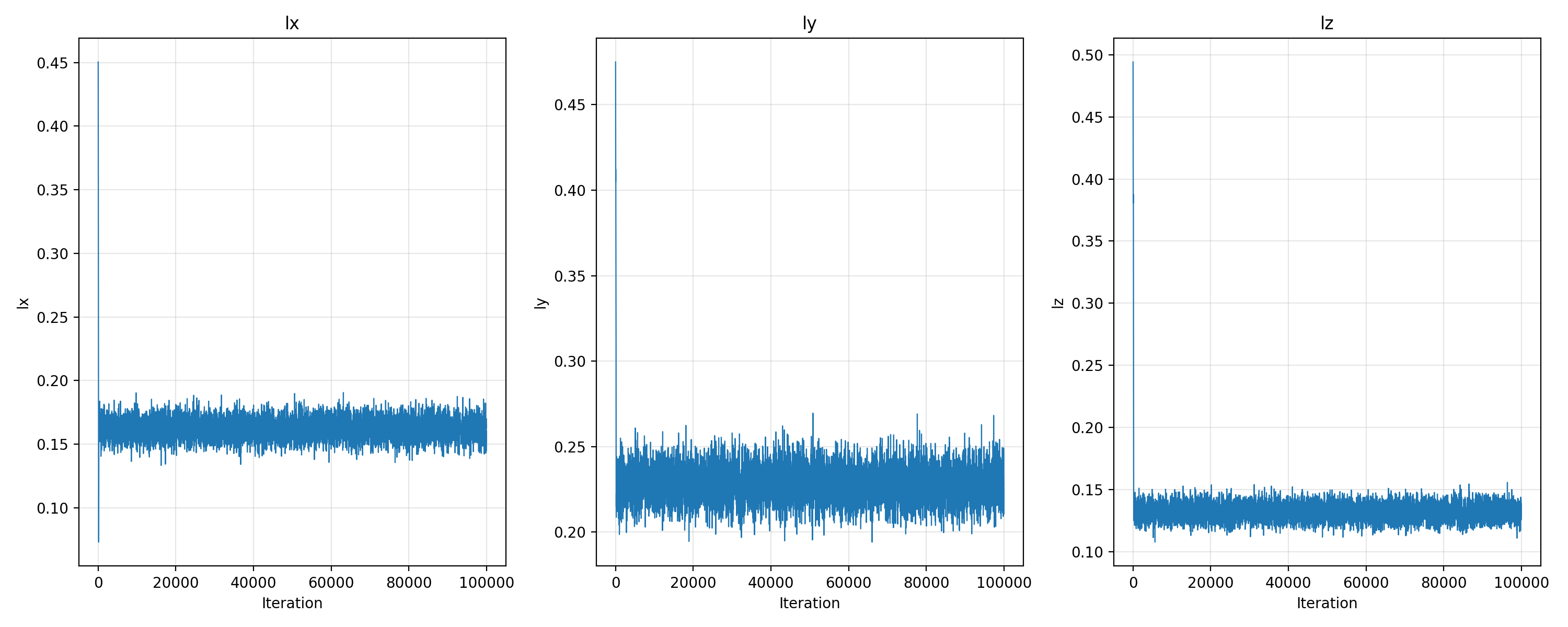}
  \end{minipage}

  \caption{MCMC traces on the rotated synthetic dataset $\bD_1$ after burn-in. Top: rotational kernel hyperparameters $(\ell_x,\ell_y,\ell_z,a_1,a_2,a_3)$. Bottom: axis-aligned ARD kernel hyperparameters $(\ell_x,\ell_y,\ell_z)$.}
  \label{fig:traces-rotdata-both}
\end{figure}

\begin{figure}[!htbp]
  \centering
  \begin{minipage}{0.65\columnwidth}
    \centering
    \includegraphics[width=\linewidth]{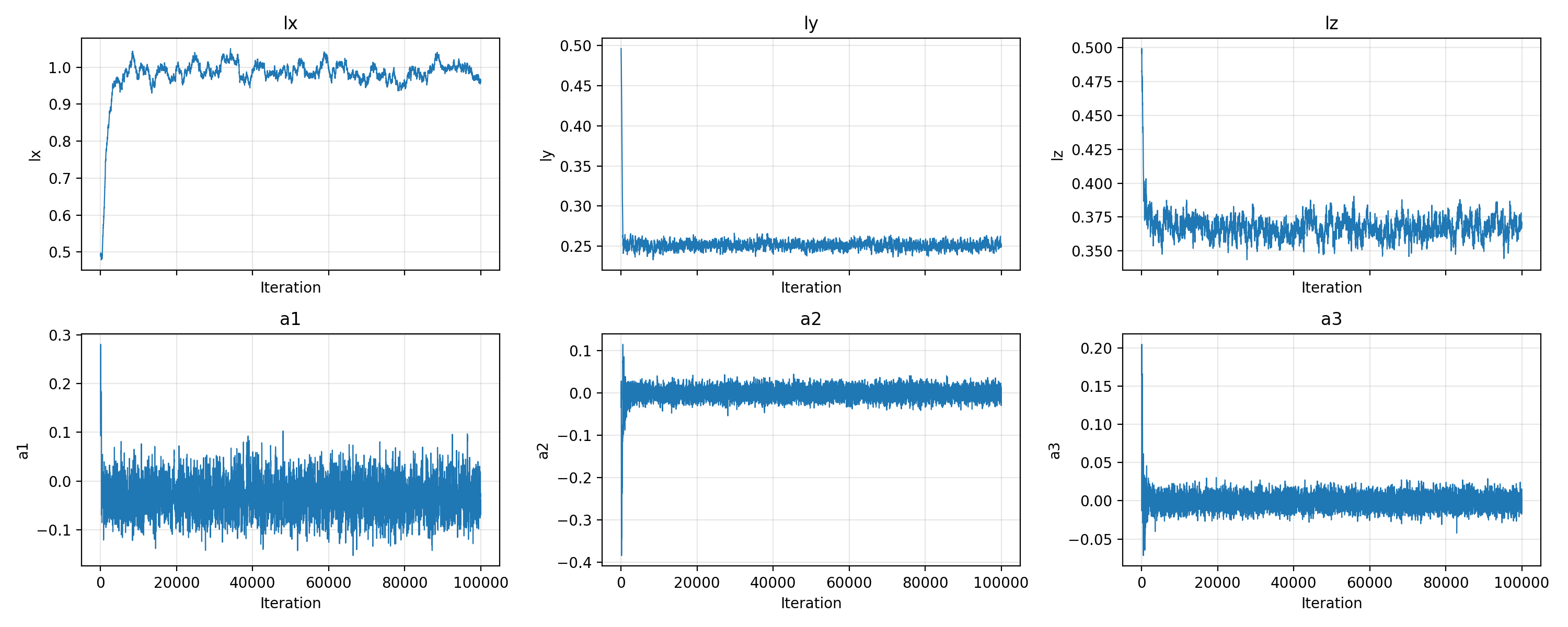}
  \end{minipage}

  \vspace{0.5em}

  \begin{minipage}{0.65\columnwidth}
    \centering
    \includegraphics[width=\linewidth]{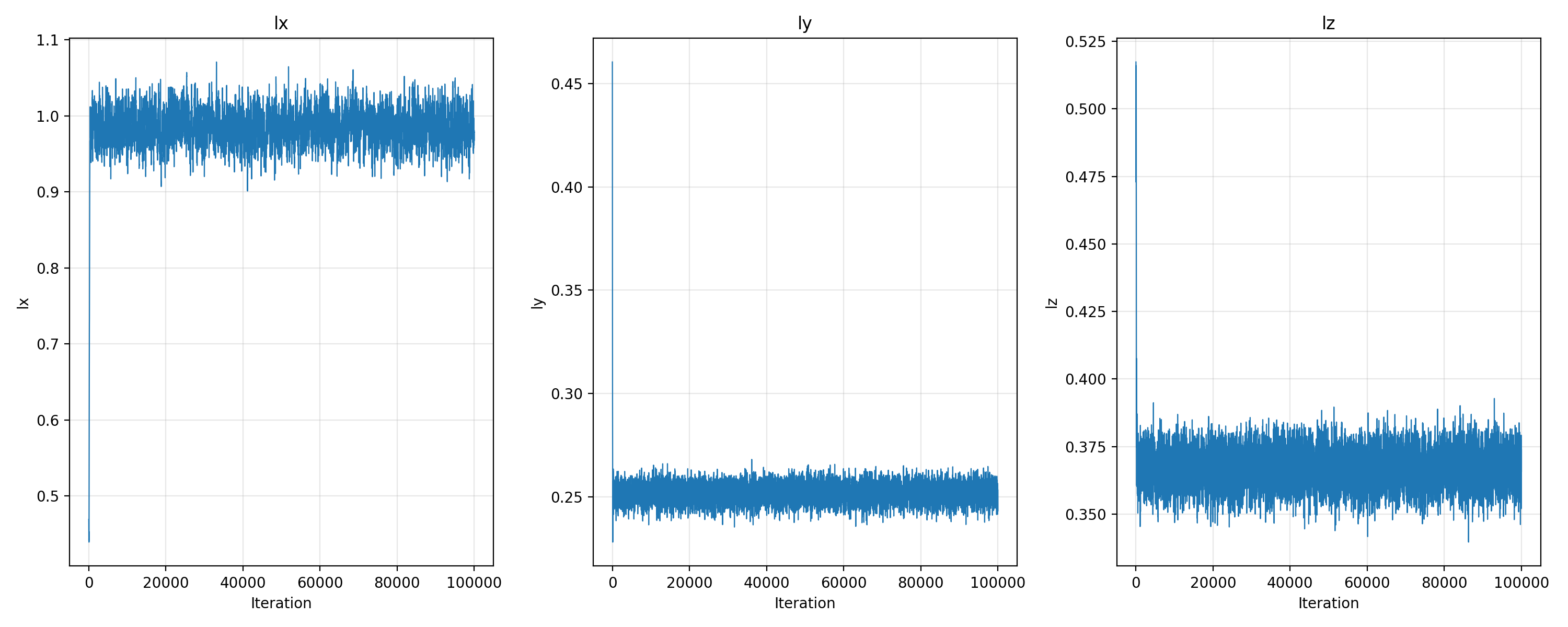}
  \end{minipage}

  \caption{MCMC traces on the axis-aligned synthetic dataset $\bD_2$ after burn-in. Top: rotational kernel hyperparameters $(\ell_x,\ell_y,\ell_z,a_1,a_2,a_3)$. Bottom: axis-aligned ARD kernel hyperparameters $(\ell_x,\ell_y,\ell_z)$.}
  \label{fig:traces-normaldat-both}
\end{figure}

We assess MCMC mixing and convergence using trace plots of the kernel hyperparameters. Figure~\ref{fig:traces-rotdata-both} shows traces for the rotated synthetic dataset $\bD_1$, and Figure~\ref{fig:traces-normaldat-both} shows the corresponding traces for the axis-aligned synthetic dataset $\bD_2$.

For $\bD_1$, the traces illustrate stable behaviour for the length-scales in both models and for the rotation parameters in the rotational model. For $\bD_2$, the rotational kernel again shows stable behaviour in the length-scales, while the rotation parameters remain close to zero. This is consistent with the hyperparameter recovery in Table~\ref{tab:synthetic-hyperparams}, where the learnt rotation is near the identity for data generated from an axis-aligned kernel. We do not plot traces for the generic SPD baseline because its six unconstrained Cholesky parameters are not directly interpretable. Instead, we assess the SPD fit via predictive performance, see Table~\ref{tab:synthetic-pred}, and via the eigendecomposition of the learnt matrix, which provides a coordinate-free summary of anisotropy.

\FloatBarrier

\FloatBarrier

\section{Real-data Application}
\label{sec:realdata}

We apply the proposed rotationally anisotropic Gaussian process to learn the material density function at any three-dimensional sub-surface location $\bS=\bs :=(x,y,z)^\top \in {\mathbb R}^3$, inside a nano-block that was fabricated in a laboratory and was imaged thereafter with scanning electron microscopy, following \cite{chakrabarty2015bayesian}, to learn the value of the material density at chosen sub-surface locations $\bs_1,\ldots, \bs_N$. Then using the training set $\bD = \{(\bs_i, \rho_i)\}_{i=1}^N$, we want to learn the material density $\rho_{\text{test}} = f(x_{\text{test}},y_{\text{test}},z_{\text{test}})$, where $\rho\in{\mathbb R}_{\geq 0}$ is realised at any test sub-surface location $\bs_{\text{test}}=(x_{\text{test}},y_{\text{test}},z_{\text{test}})^\top$.

The material sample is placed in a cuboidal grid, such that the $z=0$ plane is the surface of the sample and the $z=c$ plane---for any $c \in(0,z^{(L)}]$---is parallel to the surface. The spatial coordinates $X$ and $Y$ lie on a uniform grid with a spacing of $0.05$ nanometres (nm), such that $x_i,y_i \in \{-1+0.05k:\;k=0,1,\dots,40\}$, while the depth coordinate takes values in a separate set of non-uniform levels $z_i \in \{z^{(1)},z^{(2)},\dots,z^{(L)}\}$. Here $i\in\{1,2,\ldots,N\}$.

\subsection{Setup of two tasks}
\label{subsec:realdata-setup}

We compare the proposed rotational kernel against two baselines. The first is an axis-aligned ARD kernel that restricts the metric to be diagonal. The second is a generic SPD kernel that learns the metric $\bM\in\mathbb{S}_{++}^3$, i.e.\ the set of $3\times3$ symmetric positive definite matrices with strictly positive eigenvalues.

All models use GP-based learning and Metropolis--Hastings based inference. The kernels used in the different models differ only in the parametrisation of $\bM$. The rotational model uses $\bM(\bell,\ba)$, as discussed in Section~\ref{sec:kernel}, and the ARD model is its diagonal special case. The SPD model, on the other hand, treats $\bM$ as an SPD metric that we learn from the data.

We consider two prediction tasks. In the first task---that we refer to as Task~1---we predict at locations $(0.9, y_j,z_j)$, for $j\in\{1,2,\ldots 49\}$, after holding out the entire $x=0.9$ plane from the training set. In the second prediction task, called Task~2, we hold out the $x=-0.35$ plane from the training set---along with the $x=-0.30$ and $x=-0.40$ planes from the training set, in order to increase the gap between training and test locations---to predict at 44 locations in the $x=-0.35$ plane.

In each prediction task, we compare the material density learnt by \cite{chakrabarty2015bayesian}---which we refer to as the truth---to the material density predicted at a given test input, by checking whether the true value of the density lies within predicted intervals. Here such predicted intervals include intervals of width $\pm 1\sigma$ or $\pm 2\sigma$ on either side of the predicted expectation of the density at that test input, where the predicted variance at this test input is $\sigma^2$. Alternatively, we compute the RMSE between the predicted expectation of the output at this test input and the truth, as well as the mean absolute error (MAE) between the truth and the expectation of the predicted material density.

\subsection{Predictive performance}

Table~\ref{tab:realdata-metrics} reports predictive performance for the rotational model and the two baselines. The two tasks differ primarily in the train--test geometry. In Task~1, the held-out plane is $x=0.9$, and training data remain available on neighbouring $x$-planes, so prediction is mainly interpolative. In Task~2, the target plane $x=-0.35$ is held out together with the neighbouring planes $x=-0.30$ and $x=-0.40$, so predictions must rely on training locations farther away in the $x$ direction. Hence the orientation of spatial dependence has a larger effect in Task~2, where the rotational kernel achieves the lowest MAE.

This difference should not be interpreted as the underlying field being rotated for predictions at $x=-0.35$ but axis-aligned for predictions at $x=0.9$. Both tasks use measurements from the same material block, and in both tasks the posterior under the rotational kernel places substantial mass away from the identity rotation, as quantified below using the $\mathrm{SO}(3)$ geodesic angle. Rather, when the prediction plane is close to observed planes, several anisotropic parameterisations yield similar predictions; when the gap is larger, the learnt orientation has a larger effect.

\begin{table}[!htbp]
  \centering
  \caption{Predictive performance in Task~1 and Task~2. Coverage is the
  percentage of test points whose true value lies within
  $\hat{y}\pm k\hat{\sigma}$ for $k\in\{1,2\}$.}
  \label{tab:realdata-metrics}

  \setlength{\tabcolsep}{8pt}

  \begin{tabular}{lccc}
    \toprule
    \multicolumn{4}{c}{Task~1: held-out plane at $x=0.9$} \\
    \midrule
    Model & MAE & $\mathrm{cov}_{1\sigma}$ & $\mathrm{cov}_{2\sigma}$ \\
    \midrule
    rotational   & 0.1597 & 80.43 & 97.83 \\
    axis-aligned & 0.1599 & 82.61 & 95.65 \\
    generic SPD  & 0.1641 & 82.61 & 95.65 \\
    \bottomrule
  \end{tabular}

  \vspace{0.8em}

  \begin{tabular}{lccc}
    \toprule
    \multicolumn{4}{c}{Task~2: held-out plane at $x=-0.35$} \\
    \midrule
    Model & MAE & $\mathrm{cov}_{1\sigma}$ & $\mathrm{cov}_{2\sigma}$ \\
    \midrule
    rotational   & 0.1527 & 90.70 & 95.35 \\
    axis-aligned & 0.1641 & 88.37 & 95.35 \\
    generic SPD  & 0.1595 & 88.37 & 95.35 \\
    \bottomrule
  \end{tabular}
\end{table}
\FloatBarrier

\begin{figure}[!htbp]
  \centering
  \includegraphics[width=0.85\columnwidth]{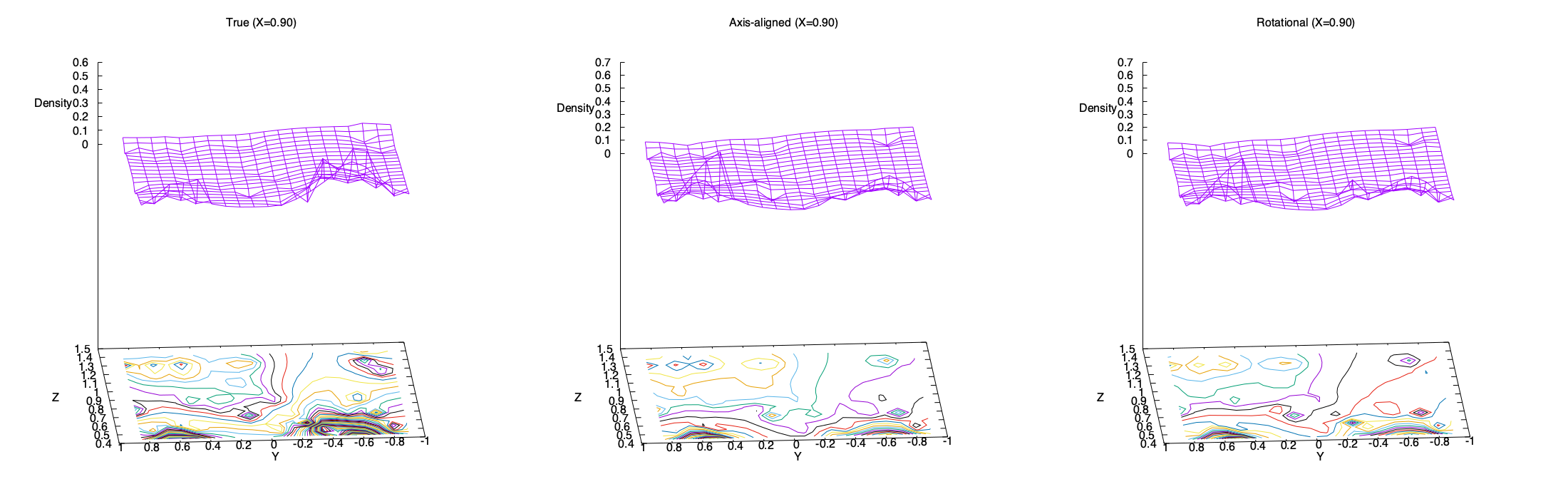}

  \vspace{-0.6em}

  \includegraphics[width=0.85\columnwidth]{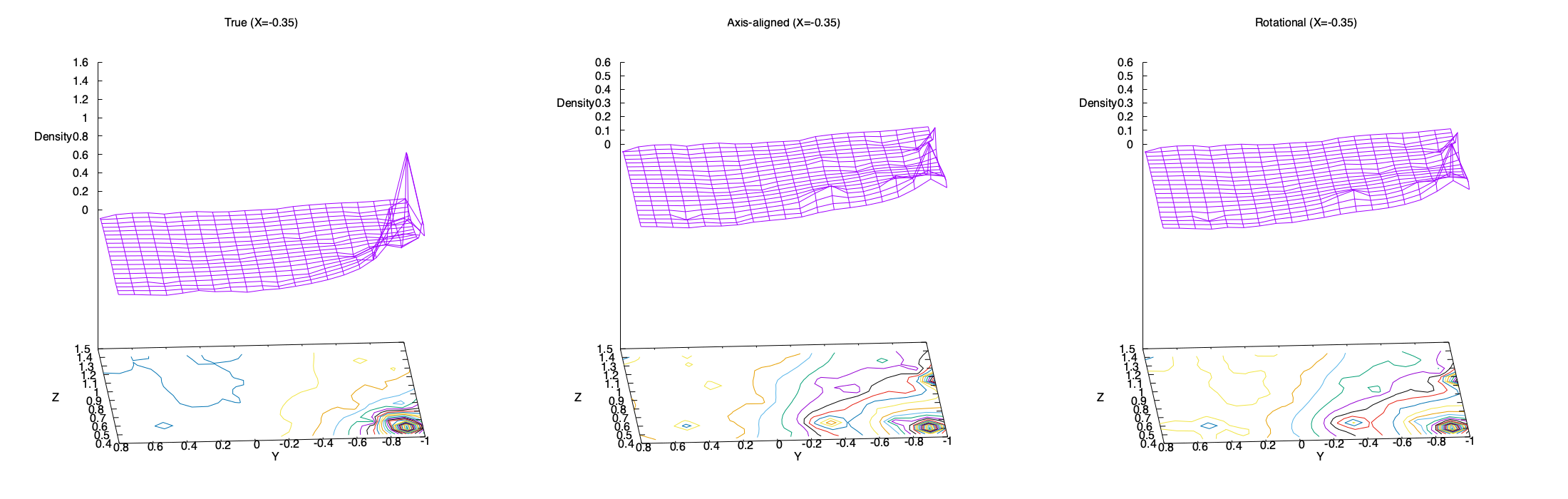}

  \vspace{-0.6em}

  \caption{True and predicted material density surfaces. Left: truth; middle: axis-aligned; right: rotational. Top: $x=0.9$; bottom: $x=-0.35$.}
  \label{fig:realdata-surfaces-both}
\end{figure}

Figure~\ref{fig:realdata-surfaces-both} plots the surface and contour plots of the material density function at the test locations in Task~1 and Task~2. The top row corresponds to the $x=0.9$ plane and the bottom row corresponds to the $x=-0.35$ plane. In each row, the true density values are shown on the left, the mean prediction under the axis-aligned kernel is shown in the middle, and the mean prediction under the rotational kernel is shown on the right. In both tasks, the predicted surfaces capture the main spatial patterns.

In Task~2, the overall shape of the true surface across the plane is matched more closely by the rotational kernel. This is again reflected in the lower MAE noted in this prediction task; see Table~\ref{tab:realdata-metrics}.

\FloatBarrier

\subsection{Learnt hyperparameters}

Table~\ref{tab:realdata-hyperparams} reports the hyperparameters of the rotational and axis-aligned kernels that we have learnt using the real-world training datasets used in the two predictive tasks, Task~1 and Task~2. For the rotational kernel we report both the principal length-scales $\bell$ and the axis--angle vector $\ba$ used to construct $\bR(\ba)\in\mathrm{SO}(3)$. The axis-aligned kernel is the diagonal special case with no rotation parameters.

\begin{table}[!htbp]
  \centering
  \caption{Learnt real-data hyperparameters. Rotation parameters are not applicable for the axis-aligned kernel.}
  \label{tab:realdata-hyperparams}
  \small
  \setlength{\tabcolsep}{7pt}

  \begin{tabular}{lrrrrrr}
    \toprule
    \multicolumn{7}{c}{Task~1} \\
    \midrule
    Model
    & $\ell_x$ & $\ell_y$ & $\ell_z$ & $a_1$ & $a_2$ & $a_3$ \\
    \midrule
    rotational
      & 0.6477 & 0.3614 & 0.03309 & 0.00998 & 0.0136 & 0.5064 \\
    axis-aligned
      & 0.4469 & 0.3341 & 0.02960 & -- & -- & -- \\
    \bottomrule
  \end{tabular}

  \vspace{0.8em}

  \begin{tabular}{lrrrrrr}
    \toprule
    \multicolumn{7}{c}{Task~2} \\
    \midrule
    Model
    & $\ell_x$ & $\ell_y$ & $\ell_z$ & $a_1$ & $a_2$ & $a_3$ \\
    \midrule
    rotational
      & 0.6872 & 0.4015 & 0.0367 & -0.0849 & 0.01005 & 0.6909 \\
    axis-aligned
      & 0.4694 & 0.3951 & 0.02777 & -- & -- & -- \\
    \bottomrule
  \end{tabular}
\end{table}
\FloatBarrier

In Task~1, the learnt rotational metric has eigenvalues $\lambda_{1}=(913.3,\;7.656,\;2.384)$ and principal ranges $\tilde{\bell}_{1}=(0.0331,\;0.361,\;0.648)$, ordered from shortest to longest. The learnt generic SPD metric has eigenvalues $\lambda_{1,\mathrm{SPD}}=(946.4,\;8.497,\;4.588)$ and principal ranges $\tilde{\bell}_{1,\mathrm{SPD}}=(0.0325,\;0.343,\;0.467)$. Both metrics are noticeably non-diagonal. The magnitude of off-diagonal structure is larger under the rotational metric in Task~1. In Task~2, the learnt rotational metric has eigenvalues $\lambda_{2}=(742.4,\;6.205,\;2.118)$ and principal ranges $\tilde{\bell}_{2}=(0.0367,\;0.4015,\;0.6872)$. The learnt generic SPD metric has eigenvalues $\lambda_{2,\mathrm{SPD}}=(972.7,\;6.902,\;3.526)$ and principal ranges $\tilde{\bell}_{2,\mathrm{SPD}}=(0.0321,\;0.381,\;0.533)$. The generic SPD kernel learns a similar qualitative pattern with one very short range and two longer ranges. Its longest range is smaller than that of the rotational fit in Task~2.

\subsection{Orientation as rotation away from axis alignment}

To summarise orientation in a coordinate-free way for the rotational model, we measure the rotation away from axis alignment using the $\mathrm{SO}(3)$ geodesic distance to the identity. For each post-burn-in sample, we compute $\bR^{(s)}=\bR(\ba^{(s)})$ and the minimal rotation angle $\theta^{(s)} = d(\bR^{(s)},\bI) = \arccos\!\left(\frac{\mathrm{tr}(\bR^{(s)})-1}{2}\right)\in[0,\pi]$.
This depends only on the induced rotation matrix and avoids interpreting raw axis--angle components directly.
In Task~1, the mean $\theta$ is learnt as $29.24^\circ$; its median is $29.7^\circ$; with a $90\%$ credible interval of $[15.97^\circ,\,40.84^\circ]$. In Task~2, the inferred rotation is larger, with a mean of $41.02^\circ$; median of $40.98^\circ$; with a $90\%$ credible interval of $[28.49^\circ,\,53.41^\circ]$. In both cases posterior mass is concentrated well away from $0^\circ$. This indicates that the dominant correlation directions are not aligned with the coordinate axes.

\subsection{Trace diagnostics}
\label{sec:realdata-traces}

\begin{figure}[!htbp]
  \centering
  \begin{minipage}{0.725\columnwidth}
    \centering
    \includegraphics[width=\linewidth]{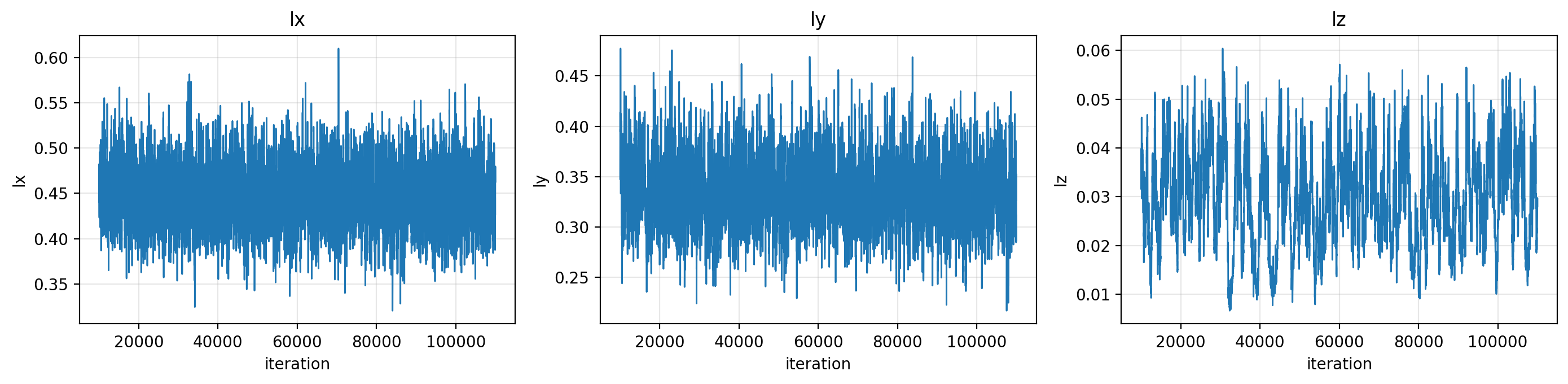}
  \end{minipage}

  \vspace{0.5em}

  \begin{minipage}{0.725\columnwidth}
    \centering
    \includegraphics[width=\linewidth]{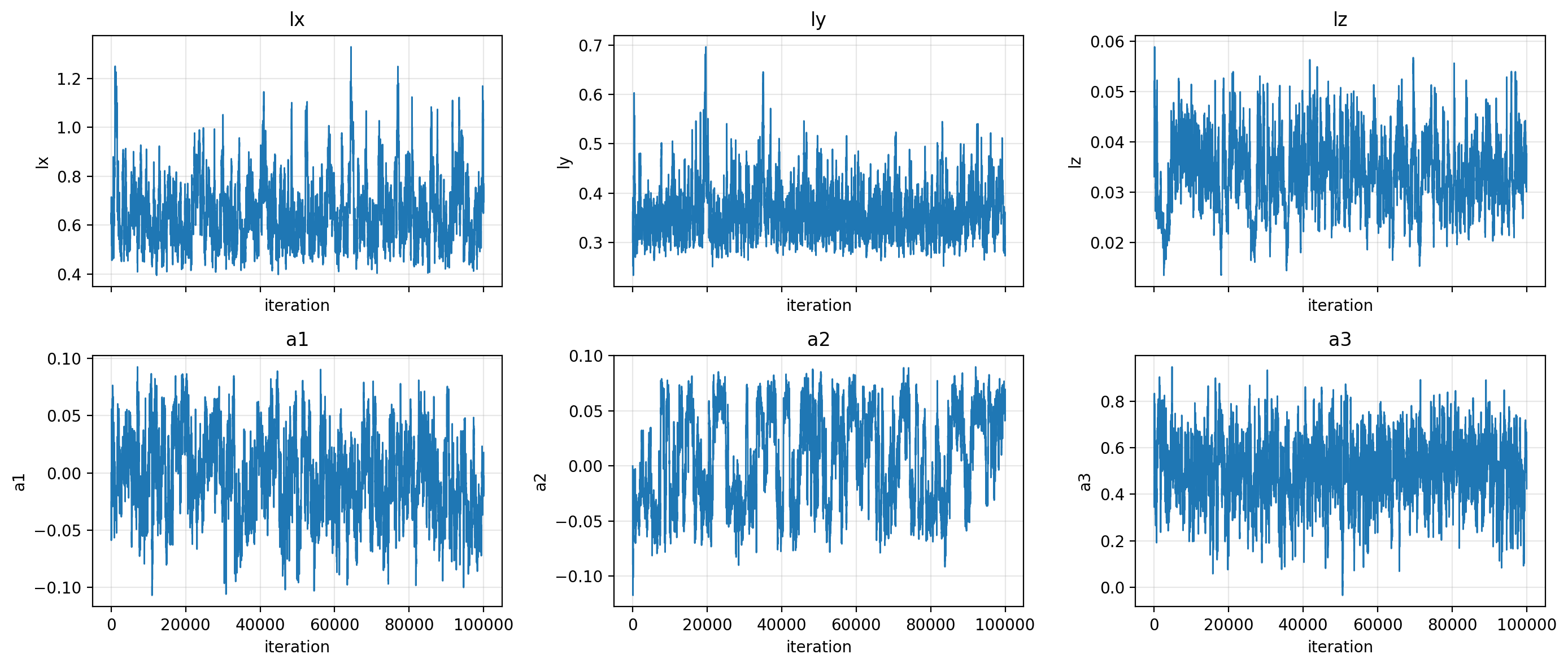}
  \end{minipage}
  \caption{Trace diagnostics for Task~1, where the held-out plane is $x=0.9$, after burn-in. Top: axis-aligned ARD kernel. Bottom: rotational kernel.}
  \label{fig:realdata-traces-x09}
\end{figure}

\begin{figure}[!htbp]
  \centering
  \begin{minipage}{0.725\columnwidth}
    \centering
    \includegraphics[width=\linewidth]{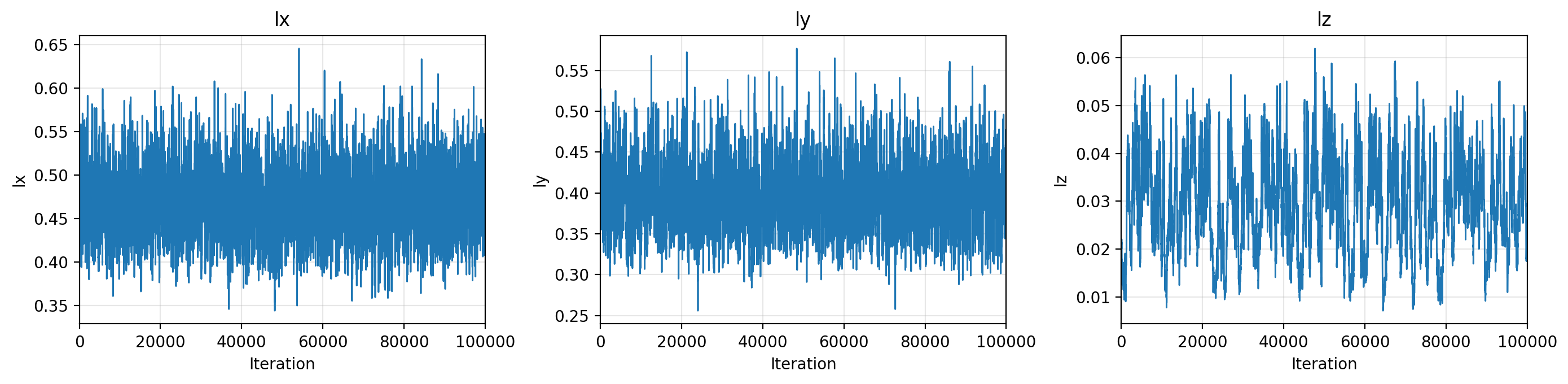}
  \end{minipage}

  \vspace{0.5em}

  \begin{minipage}{0.725\columnwidth}
    \centering
    \includegraphics[width=\linewidth]{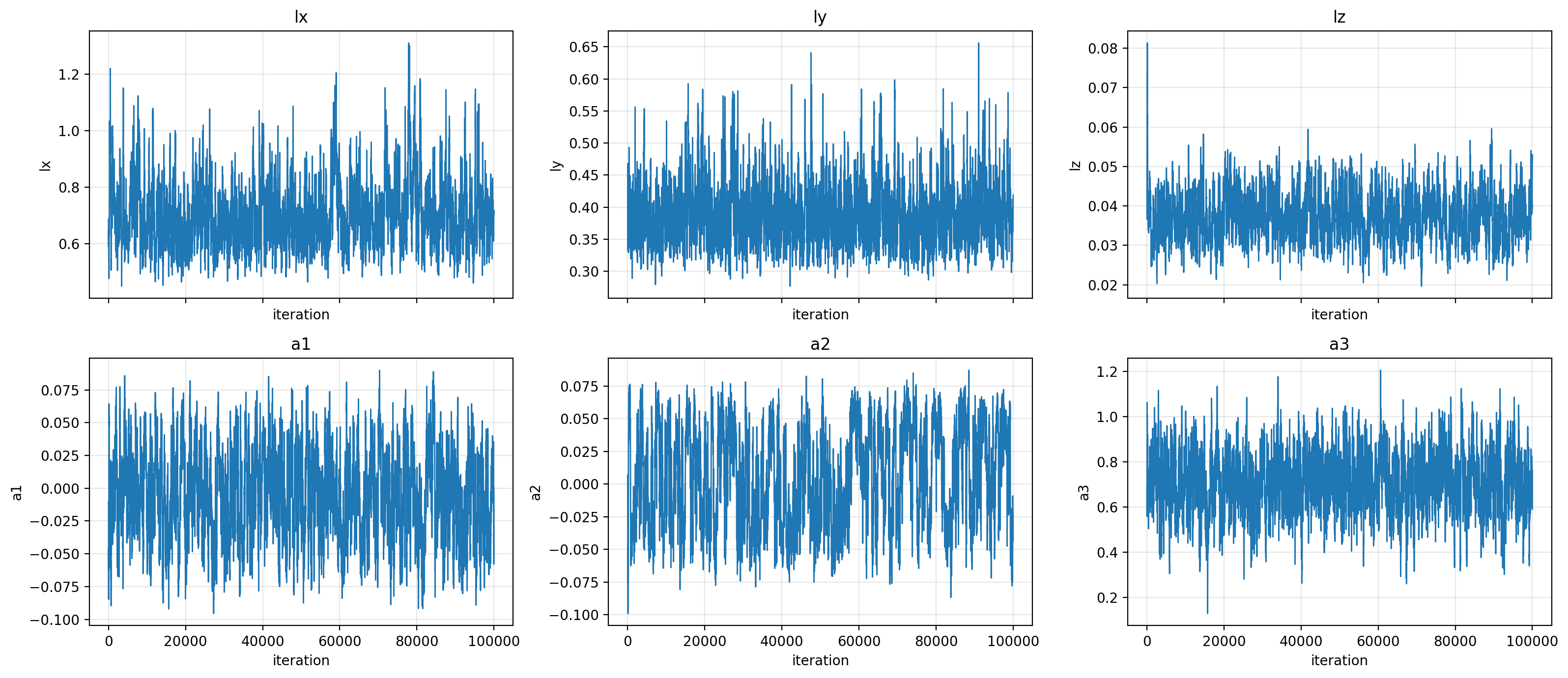}
  \end{minipage}
  \caption{Trace diagnostics for Task~2, where the held-out plane is $x=-0.35$, after burn-in. Top: axis-aligned ARD kernel. Bottom: rotational kernel.}
  \label{fig:realdata-traces-xm035}
\end{figure}

We assess MCMC mixing and convergence for both real-data prediction tasks using trace plots of the learnt kernel hyperparameters. Figure~\ref{fig:realdata-traces-x09} shows the traces for Task~1, where the held-out plane is $x=0.9$, and Figure~\ref{fig:realdata-traces-xm035} shows the traces for Task~2, where the held-out plane is $x=-0.35$.

In both tasks, the axis-aligned kernel exhibits stable mixing for the three length-scales. Under the rotational kernel, the length-scales remain well-behaved, while individual components of $\ba$ can fluctuate and change sign. This is expected given the discrete symmetries of the induced metric and the fact that the likelihood depends on $\ba$ only through $\bR(\ba)$ and hence $\bM(\bell,\ba)$. The geodesic summaries above confirm that, despite coordinate-level fluctuations, the posterior prefers a nontrivial rotation away from axis alignment in both tasks.

\FloatBarrier

\FloatBarrier

For the generic SPD model, we do not present traces of the raw parameterisation because its components do not have a clear, direct interpretation in terms of principal ranges and orientation. Instead, we summarise the fitted metric using quantities that are directly comparable across iterations, including selected entries of $\bM$ and the principal ranges obtained from its eigendecomposition. In our runs these derived summaries are stable and indicate a non-diagonal metric in both splits.

\subsection{Random multi-plane hold-out experiment}
\label{subsec:realdata-randomplanes}

We perform an additional robustness check on the real material block by holding out multiple $x$-planes selected at random and predicting on all held-out planes using a single model fit on the remaining data. Specifically, we randomly hold out the five planes
$x\in\{-0.25,\;0.10,\;0.15,\;0.60,\;0.85\}$ and evaluate predictions on all test locations across these planes. We report mean absolute error (MAE) and empirical coverage of nominal $\pm1\sigma$ and $\pm2\sigma$ predictive intervals.

This learning scheme differs from the two prediction tasks in Section~\ref{subsec:realdata-setup} in two important ways. First, several $x$-planes are held out at once, so performance is assessed across multiple cross-sections simultaneously. Second, since the held-out planes come from within the block, the task focuses on generalisation across slices of the same material rather than across one larger missing region. We use this setting as a robustness check for whether the rotational parameterisation delivers consistent gains across multiple planes at once.

\begin{figure}[!htbp]
  \centering
  \includegraphics[width=0.85\columnwidth]{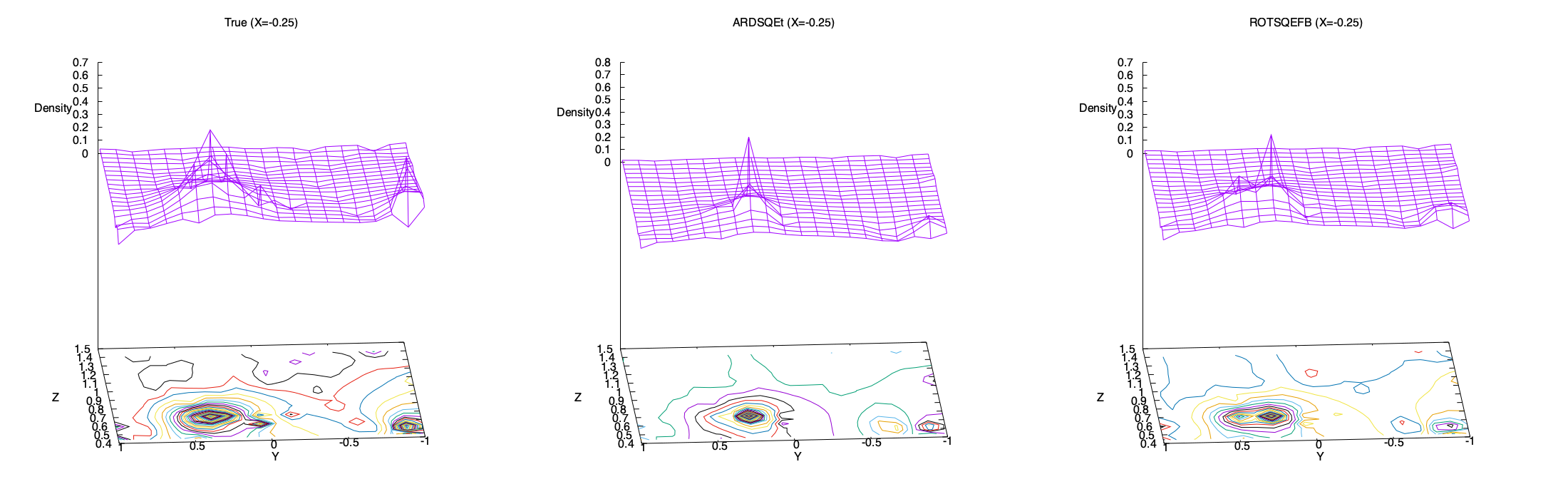}
  \caption{Surface and contour plots at test locations in the $x=-0.25$ plane. Left: true density. Middle: axis-aligned kernel. Right: rotational kernel.}
  \label{fig:multiheld-surfaces-xm025}
\end{figure}

\begin{figure}[!htbp]
  \centering
  \includegraphics[width=0.85\columnwidth]{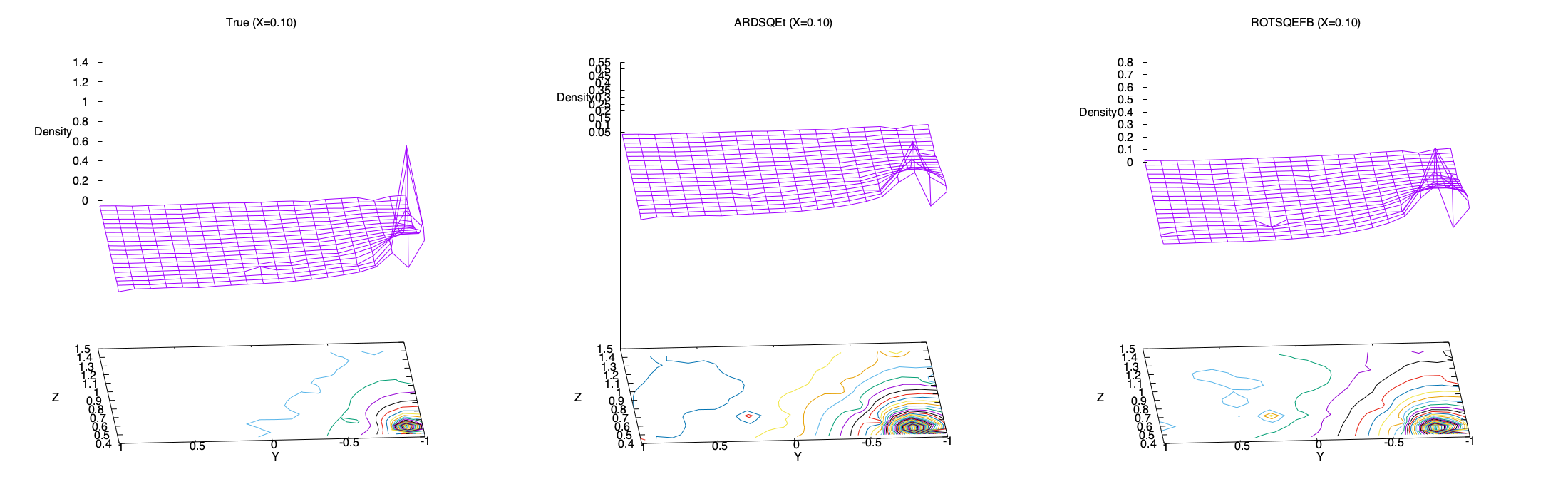}
  \caption{Same as Figure~\ref{fig:multiheld-surfaces-xm025}, except the test locations are on the $x=0.10$ plane.}
  \label{fig:multiheld-surfaces-x010}
\end{figure}

\begin{figure}[!htbp]
  \centering
  \includegraphics[width=0.85\columnwidth]{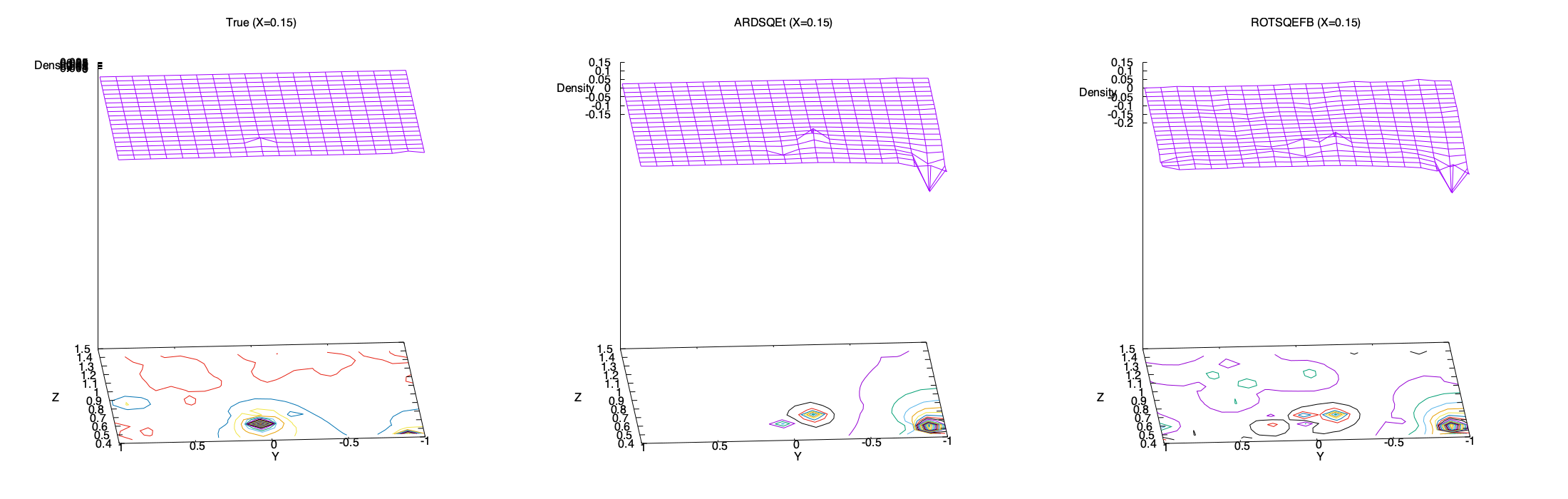}
  \caption{Same as Figure~\ref{fig:multiheld-surfaces-xm025}, except the test locations are on the $x=0.15$ plane.}
  \label{fig:multiheld-surfaces-x015}
\end{figure}

\begin{figure}[!htbp]
  \centering
  \includegraphics[width=0.85\columnwidth]{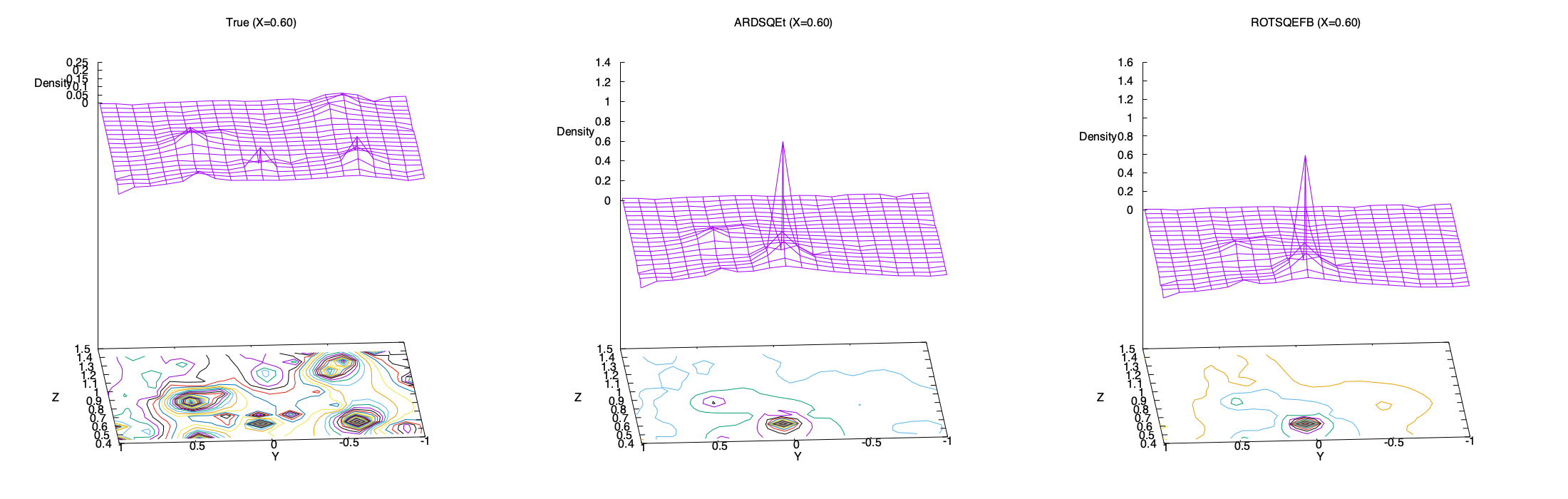}
  \caption{Same as Figure~\ref{fig:multiheld-surfaces-xm025}, except the test locations are on the $x=0.60$ plane.}
  \label{fig:multiheld-surfaces-x060}
\end{figure}

\begin{figure}[!htbp]
  \centering
  \includegraphics[width=0.85\columnwidth]{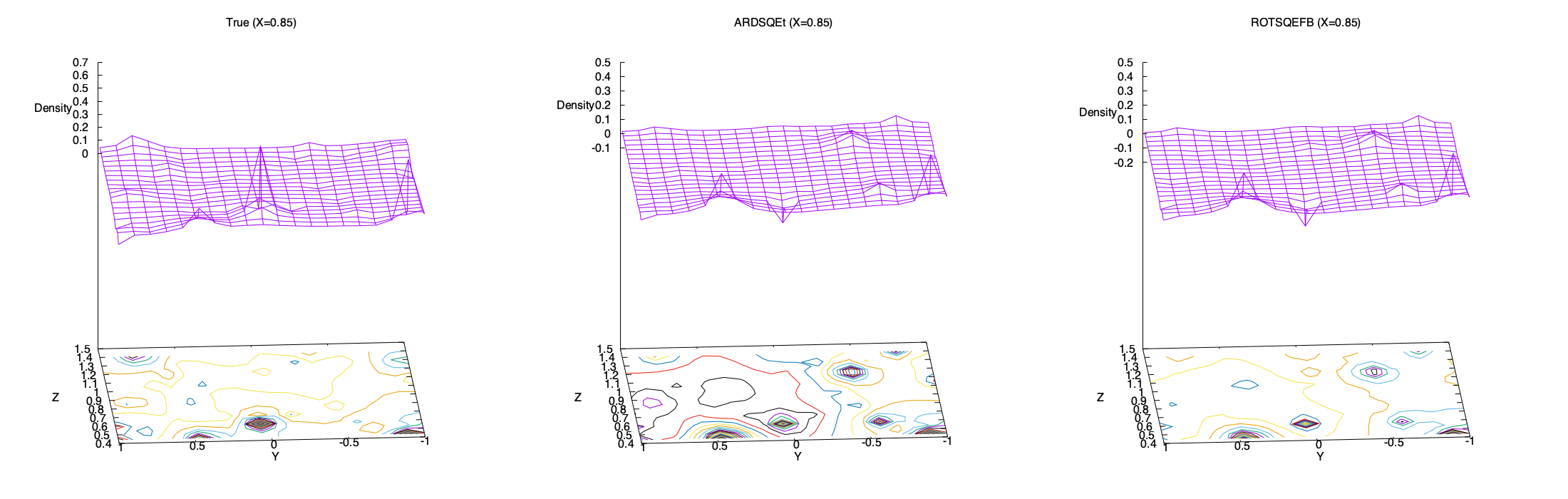}
  \caption{Same as Figure~\ref{fig:multiheld-surfaces-xm025}, except the test locations are on the $x=0.85$ plane.}
  \label{fig:multiheld-surfaces-x085}
\end{figure}

Table~\ref{tab:randomplanes-metrics} reports predictive performance computed over all evaluated held-out planes, with $N=245$ test points. For clarity, the visualisations in Figures~\ref{fig:multiheld-surfaces-xm025}--\ref{fig:multiheld-surfaces-x085} compare only the axis-aligned ARD and rotational kernels; the generic SPD baseline is included in Table~\ref{tab:randomplanes-metrics} for completeness. In this random multi-plane setting, the rotational kernel achieves the lowest MAE. Coverage remains comparable across models: $2\sigma$ coverage is essentially unchanged relative to the axis-aligned baseline, while $1\sigma$ coverage is slightly lower. Overall, the improvement is reflected primarily in reduced absolute error rather than in wider predictive intervals.

\begin{table}[!htbp]
  \centering
  \caption{Predictive performance for the random multi-plane hold-out experiment on the real material block. Coverage is the percentage of test points whose true value lies within $\hat{y}\pm k\hat{\sigma}$ for $k\in\{1,2\}$.}
  \label{tab:randomplanes-metrics}

  \begin{tabular}{lccc}
    \toprule
    Model & MAE & $\mathrm{cov}_{1\sigma}(\%)$ & $\mathrm{cov}_{2\sigma}(\%)$ \\
    \midrule
    rotational   & 0.110874 & 94.29 & 97.55 \\
    axis-aligned & 0.122082 & 95.10 & 97.14 \\
    generic SPD  & 0.115637 & 92.65 & 97.14 \\
    \bottomrule
  \end{tabular}
\end{table}
\FloatBarrier

To assess whether the aggregate improvement is driven by a single held-out plane or is consistent across slices, Table~\ref{tab:randomplanes-mae-byx} reports per-plane MAE. The rotational kernel achieves lower MAE than the axis-aligned kernel on each of the five held-out planes, showing that the aggregate gain is not dominated by one favourable cross-section. Differences in effect size across planes are expected, since the local structure of the field varies from slice to slice.

\begin{table}[!htbp]
  \centering
  \caption{Per-plane MAE for the random multi-plane prediction experiment. Each held-out plane contains $N=49$ test points.}
  \label{tab:randomplanes-mae-byx}

  \begin{tabular}{lcc}
    \toprule
    Prediction plane $x$ & rotational & axis-aligned \\
    \midrule
    $x=-0.25$ & 0.1341 & 0.1460 \\
    $x=0.10$  & 0.1637 & 0.1731 \\
    $x=0.15$  & 0.0697 & 0.0897 \\
    $x=0.60$  & 0.0883 & 0.0984 \\
    $x=0.85$  & 0.0986 & 0.1032 \\
    \bottomrule
  \end{tabular}
\end{table}
\FloatBarrier

Figures~\ref{fig:multiheld-surfaces-xm025}--\ref{fig:multiheld-surfaces-x085} plot surface and contour plots of the material density function at test locations in each held-out plane. In each figure, the left panel shows the true density, the middle panel shows predictions under the axis-aligned kernel, and the right panel shows predictions under the rotational kernel.

\FloatBarrier
\FloatBarrier
\FloatBarrier
\FloatBarrier
\FloatBarrier
\section{Conclusion}

This paper developed a rotationally anisotropic Gaussian process kernel for three-dimensional spatial fields, with the aim of making the fitted covariance geometry directly interpretable. The model represents anisotropy through principal correlation ranges together with an explicit orientation in $\mathrm{SO}(3)$. Using axis--angle coordinates and the Lie-algebra exponential map gives an unconstrained parameterisation for inference while ensuring that every parameter value induces a valid SPD covariance metric.

The contribution is not an expansion of the covariance family beyond generic full-SPD metrics, but a change in what the model treats as primitive. In a generic SPD parameterisation, the scientifically relevant quantities, that being the principal ranges and directions, are recovered only indirectly from the fitted matrix. Here, they are the parameters on which inference is performed. This allows priors, proposals, and posterior uncertainty to be expressed directly in terms of the principal ranges $\bell$ and orientation $\ba$, rather than through unconstrained matrix entries whose geometric meaning is only available after eigendecomposition. It also makes the non-identifiabilities of the representation explicit: different values of $(\bell,\ba)$ may induce the same metric $\bM$, and rotations become weakly identified when principal ranges are nearly equal.

The synthetic experiments show that this parameterisation behaves as intended. When the data are generated from a rotated anisotropic covariance, the model recovers the underlying metric and gives substantially better predictions than an axis-aligned ARD kernel. Its predictive accuracy matches that of a generic SPD baseline, as expected because the two models span the same metric family, while the rotational parameterisation yields posterior summaries directly in geometric terms. When the data are generated from an axis-aligned covariance, the inferred rotation remains near the identity and predictive performance matches ARD, indicating that the additional rotational degrees of freedom do not harm performance when they are unnecessary.

On the material-density dataset, the learned covariance geometry is consistently non-axis-aligned. The inferred rotations away from the identity, together with the estimated principal ranges, provide a compact description of the dominant directions of spatial dependence in the three-dimensional block. These summaries reveal anisotropic structure that is not represented by an axis-aligned kernel. Overall, the proposed kernel provides a way to learn spatial dependence in a form that is statistically flexible, geometrically meaningful, and directly interpretable.

\bibliographystyle{ACM-Reference-Format}
\bibliography{references}

\end{document}